\documentclass[10pt]{article}
\usepackage[preprint]{tmlr}

\usepackage{amsmath,amsfonts,bm}

\def\eqref#1{equation~\ref{#1}}

\def\1{\bm{1}}

\DeclareMathAlphabet{\mathsfit}{\encodingdefault}{\sfdefault}{m}{sl}
\SetMathAlphabet{\mathsfit}{bold}{\encodingdefault}{\sfdefault}{bx}{n}

\usepackage{hyperref}
\usepackage{url}
\usepackage{graphicx}
\usepackage{multirow}
\usepackage{tabularx}
\usepackage{booktabs}
\usepackage{array}
\usepackage{placeins}
\usepackage{pifont}
\usepackage{mathtools}
\usepackage{amsmath,amssymb,bbold}
\usepackage[table]{xcolor}
\usepackage{wrapfig}
\usepackage{caption}
\usepackage{subcaption}
\usepackage{rotating}

\newcommand{\eightpt}{\fontsize{8}{9.5}\selectfont}
\newcommand{\parhead}[1]{\par\smallskip\noindent\textbf{#1}\ }
\newcommand{\cmark}{\textcolor{green!80!lime}{\ding{51}}}
\newcommand{\xmark}{\textcolor{red!80!orange}{\ding{55}}}
\newcommand{\PaperVersion}{2}

\title{CURE-OOD: Benchmarking Out-of-Distribution Detection for Survival Prediction}

\author{Wenjie Zhao$^{1}$ \hspace{0.5cm} Jia Li$^{1}$ \hspace{0.5cm} Mingrui Liu$^{2}$ \hspace{0.5cm} Jing Wang$^{3}$ \hspace{0.5cm} Yunhui Guo$^{1}$\\
{\small\itshape
$^{1}$University of Texas at Dallas \quad
$^{2}$George Mason University \quad
$^{3}$UT Southwestern Medical Center}\\[0.5em]
{\small
$^{1}$\texttt{\{wenjie.zhao, yunhui.guo\}@utdallas.edu}
}
}

\def\month{04}
\def\year{2026}
\def\openreview{\url{https://openreview.net/forum?id=TBD}}

\begin{document}

\maketitle

\begin{abstract}
``How long can I live and remain free of cancer?''
is often the first question a patient asks after receiving a cancer diagnosis and treatment.
Accurate survival prediction helps alleviate psychological distress and supports risk stratification and personalized treatment planning.
Recent survival prediction frameworks have shown strong performance using computed tomography (CT) images.
However, variations in imaging acquisition introduce out-of-distribution (OOD) samples caused by covariate shifts that undermine model reliability.
Despite this challenge, to our knowledge, no existing benchmark systematically studies OOD detection in cancer survival prediction.
To address this gap, we introduce the Cancer sURvival bEnchmark for OOD Detection (CURE-OOD), the first benchmark for systematically evaluating OOD detection in survival prediction under controlled acquisition-induced distribution shifts.
CURE-OOD defines scanner-parameter-based training, in-distribution (ID), and OOD test splits across four survival prediction tasks.
Our experiments show that covariate shifts notably reduce survival prediction performance.
It also shows that mainstream classification-oriented OOD detectors can fail in survival prediction.
Finally, we include HazardDev as a simple survival-aware reference baseline for OOD detection.
CURE-OOD enables systematic analysis of how distribution shifts affect both downstream survival performance and OOD detectability.
\end{abstract}

\section{Introduction}
\label{sec:intro}

Accurate survival prediction is critical in cancer prognosis, as it helps clinicians assess patient risk and make informed, personalized treatment decisions.
In head and neck cancer (HNC), radiation therapy is typically guided by stage-dependent risk stratification protocols, where patients within the same stage receive identical dose prescriptions \citep{PAN2016493, CAUDELL2017e266}.
Yet patients with similar staging or tumor characteristics can experience markedly different outcomes, ranging from long-term recurrence-free survival to early disease progression or treatment-related toxicity \citep{Ang201024, kawecki2014follow}.
This heterogeneity motivates reliable, individualized survival prediction models that complement traditional staging systems and guide treatment planning \citep{beesley2019individualized, KANG20151127, amin2017ajcc}.

\begin{figure*}[t]
    \centering
    \begin{minipage}[t]{0.49\textwidth}
        \centering
        \includegraphics[width=\linewidth]{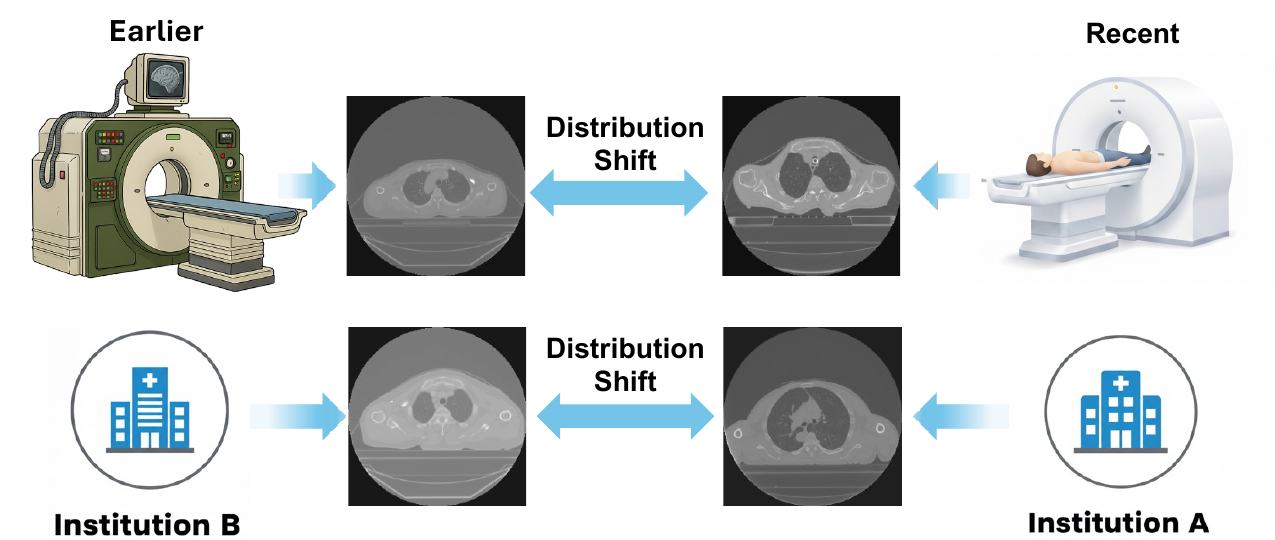}
        \captionof{figure}{CT data distribution shifts caused by cross-institution variability and scanner changes over time.}
        \label{fig:shift_source}
    \end{minipage}\hfill
    \begin{minipage}[t]{0.49\textwidth}
        \centering
        \includegraphics[width=\linewidth]{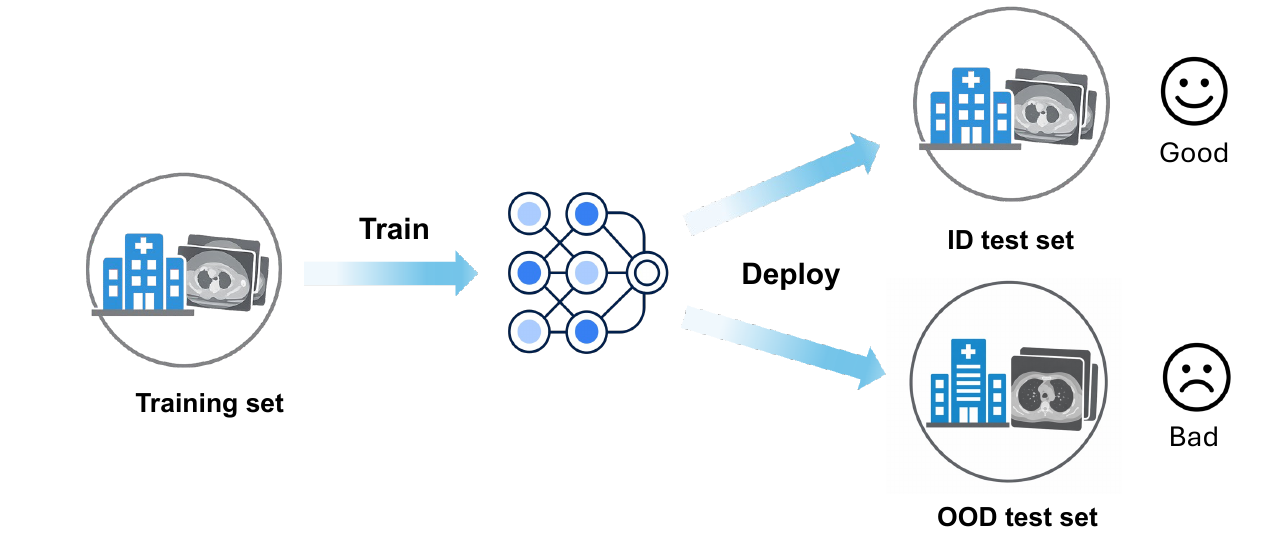}
        \captionof{figure}{A model trained on ID data performs well on ID test data but degrades on OOD test data.}
        \label{fig:id_ood_gap}
    \end{minipage}
\end{figure*}

Recent advances in deep learning have enabled survival modeling from medical images.
Typically, CNNs or Vision Transformers (ViTs) extract imaging features that are passed to survival prediction frameworks \citep{chen2024vision, saeed2024survrnc, chen2024advancing}.
A representative formulation is multi-task logistic regression (MTLR), which reformulates survival analysis into ordered binary subtasks across discrete time intervals to jointly learn a survival distribution \citep{zhang2021survey, yu2011learning, kvamme2019time, nagpal2021deep}. In real-world practice, CT scans are often collected over long periods and across hospitals or imaging centers.
During this process, scanner hardware and acquisition protocols may be updated or replaced \citep{welch2023computed}, while cross-center variation further introduces differences in scanner models and acquisition settings \citep{guan2021domain}, as illustrated in Fig.~\ref{fig:shift_source}.
This variability changes image appearance and induces covariate shifts while the clinical prediction task remains fixed.
In image classification, segmentation, and related medical imaging tasks, such shifts have been shown to reduce model reliability \citep{gutbrod2025openmibood, liu2025h2st, baek2024unexplored, zhao2025pixel, zhao2026mitigating}.
However, for cancer survival prediction, whether acquisition-induced covariate shifts cause significant downstream performance degradation has not been systematically studied.


Existing OOD benchmarks and detection methods mainly target classification, where detectors often rely on class-confidence or logit-based signals. Survival prediction differs from classification because its outputs are time-indexed risk or survival distributions rather than class confidence. Therefore, whether classification-oriented OOD detectors remain effective for survival prediction under acquisition-induced shifts has not been systematically studied.

To address these gaps, we introduce CURE-OOD, a benchmark for OOD detection in cancer survival prediction.
Table~\ref{tab:benchmarks_compare} summarizes key differences between CURE-OOD and existing OOD benchmarks.
Following standard OOD terminology, in-distribution (ID) data are drawn from the same distribution as the training set, whereas OOD data deviate from that distribution.
We build CURE-OOD as a controlled evaluation protocol with non-overlapping training, in-distribution (ID) test, and OOD test splits.
These splits are defined by four acquisition parameters, including pixel spacing, exposure time, slice thickness, and X-ray tube current.
This protocol enables systematic and reproducible evaluation of two questions.
First, whether acquisition-induced covariate shifts degrade survival prediction.
Second, whether existing OOD detection methods can reliably identify shifted samples in survival modeling.
On CURE-OOD, we find that these shifts reduce survival prediction performance and that several classification-oriented OOD detectors fail under survival formulations, sometimes even assigning more OOD-like scores to ID samples.
We further include HazardDev as a simple survival-aware reference baseline that uses predicted hazard as the detection signal.

Our main contributions are summarized as follows:
\begin{itemize}
    \item We present CURE-OOD, to our knowledge, the first benchmark for out-of-distribution detection in cancer survival prediction built on real-world clinical CT data, extending medical imaging OOD evaluation beyond classification and segmentation.
    \item We systematically verify the degradation of survival prediction performance under acquisition-induced covariate shifts and benchmark representative OOD detectors in this setting.
    \item We introduce HazardDev, a simple survival-aware OOD detection baseline based on predicted hazard, and analyze the limitations of existing classification-oriented detectors for survival prediction.
\end{itemize}


CURE-OOD facilitates more rigorous evaluation of OOD detection in survival prediction and supports the development of survival-aware methods under acquisition-induced shifts.
\section{Related work}
\label{sec:related_work}

\noindent\textbf{Survival prediction in medical imaging.}
Modern survival modeling has evolved from classical statistics to deep learning. Early work relied on the Cox proportional hazards model \citep{cox1972regression} and Aalen regression models \citep{aalen1989linear}, which are powerful for statistical inference but limited for individualized prediction and rely on the proportional hazards assumption. DeepSurv \citep{katzman2018deepsurv} and other “deep-Cox” approaches \citep{martinussen2006dynamic, ching2018cox, nagpal2021deep, kvamme2019time} extend these hazard-based formulations while using neural networks for representation learning. In contrast, MTLR \citep{yu2011learning} discretizes time and learns a sequence of dependent logistic classifiers, directly modeling the survival function and naturally handling censored data. This reformulation makes survival prediction operationally similar to classification across time intervals. In current imaging pipelines, MTLR is commonly combined with visual backbones such as CNNs or ViTs to extract features and predict patient-specific risk or hazard over time \citep{wiegrebe2024deep}. It has also been adopted in recent multilabel survival prediction systems \citep{kim2021deep}.

\noindent\textbf{Covariate shift in medical datasets.}
Medical imaging datasets are often collected over many years, especially for survival prediction tasks requiring decade-long follow-up \citep{welch2024radcure, henschke202320, littlejohns2020uk}.
During this span, scanner hardware, protocols and reconstruction algorithms are frequently upgraded or replaced to meet evolving clinical and technical standards \citep{fortin2018harmonization}. In addition, multi-center studies introduce further variability across institutions and device vendors, as identical scanners may operate under different acquisition conditions \citep{mackin2015measuring}.
These factors yield heterogeneous acquisition settings that alter image appearance and quantitative features, leading to covariate shift and degraded downstream performance \citep{zech2018variable}. 
Unlike many computer vision applications, medical imaging demands much higher reliability and safety, where incorrect predictions can directly impact clinical decisions and patient outcomes. Therefore, rather than solely adapting models to unseen domains, it is crucial to detect OOD samples that arise from covariate shifts \citep{cao2020benchmark, gutbrod2025openmibood}, allowing clinicians to recognize potentially unreliable predictions before they affect patient care.

\noindent\textbf{OOD detection.}
OOD detection aims to identify inputs that deviate from the training distribution and has been widely studied in computer vision. Most existing methods are designed for image classification datasets such as CIFAR and ImageNet, relying on confidence- or representation-based scores, including maximum softmax probability (MSP) \citep{hendrycks2016baseline}, ODIN \citep{liang2017enhancing}, energy-based scoring \citep{liu2020energy} and activation shaping (ASH) \citep{djurisic2022extremely}. Some studies further extend OOD detection to segmentation tasks, focusing on pixel-level anomaly localization \citep{zhao2025pixel, zhao2024segment, vojivr2024pixood, shoeb2024have}.
In the medical domain, Cao et al. \citep{cao2020benchmark} provided one of the earliest systematic evaluations of OOD detection across several medical imaging modalities. More recently, Gutbrod et al. introduced OpenMIBOOD \citep{gutbrod2025openmibood}, a large-scale benchmark inspired by OpenOOD \citep{yang2022openood} that comprehensively assesses OOD detection under medical image classification settings.
Despite considerable progress, OOD detection has been explored mainly in classification and segmentation tasks. To our knowledge, survival prediction, with its distinct time-to-event objectives and censoring-aware formulations, has not yet been systematically studied.

\ifnum\PaperVersion=1
    \section{Background}
\label{sec:bg}

As illustrated in Fig.~\ref{fig:shift_source}, the distribution of medical imaging data is shaped by multiple factors along the acquisition pipeline. 
These include the physical properties of the tissue being scanned, the scanning environment, and most importantly, the configuration of the imaging device. 
In CT imaging, scanner parameters such as pixel spacing, exposure time, slice thickness, and X-ray tube current play a central role in defining spatial resolution, image contrast, and noise characteristics. 
Variations in these acquisition parameters may result from differences in hardware models, calibration protocols, or clinical requirements across institutions and time periods. 
This section provides the background for understanding our benchmark. 
We first outline the survival prediction task and its commonly used modeling framework, then describe the CT acquisition process and discuss how parameter variability introduces covariate shifts that affect model reliability.


\begin{table*}[t]
\centering
\small
\setlength{\tabcolsep}{4pt}
\renewcommand{\arraystretch}{1.05}
\begin{tabularx}{\linewidth}{l c c c c}
\toprule
\textbf{Benchmark / Dataset} & \textbf{Medical} & \textbf{Primary Task} & \textbf{Shift Type} & \textbf{Data Source} \\
\midrule
OpenOOD \citep{yang2022openood}& \xmark & Classification & Semantic (near/far) & Mixed (real + synthetic) \\
OpenMIBOOD \citep{gutbrod2025openmibood}& \cmark & Classification / Segmentation & Covariate + Semantic & Real (medical scans) \\
ImageNet-ES \citep{baek2024unexplored}& \xmark & Classification & Covariate (Env./Sensor) & Real (camera images) \\
ImageNet-C / CIFAR-C \citep{hendrycks2019benchmarking}& \xmark & Classification & Covariate (Synthetic corruption) & Synthetic (noise / blur) \\
ImageNet-A / -O / -R \citep{hendrycks2021natural}& \xmark & Classification & Semantic (Hard/Open/Style) & Real (collected subsets) \\
WILDS \citep{pmlrv139koh21a} & Partially  & Classification / Regression & Covariate (Domain/Site/Batch) & Real (multi-domain) \\
MIDOG \citep{AUBREVILLE2023102699} & \cmark & Detection & Covariate (Scanner/Stain) & Real (histopathology) \\
PhaKIR \citep{rueckert2024phakir} & \cmark & Detection / Classification & Covariate (Institution/Camera) & Real (surgical videos) \\
OASIS-3 \citep{lamontagne2019oasis} & \cmark & Classification & Covariate (Site/Scanner) & Real (MRI data) \\
MedMNIST v2 \citep{yang2023medmnist} & \cmark & Classification & Mixed (Dataset-level) & Mixed (real + curated) \\
BREEDS / Living-17 \citep{santurkar2020breeds} & \xmark & Classification & Subpopulation shift & Real (natural images) \\
\textbf{CURE-OOD} & \cmark & Risk modeling & Covariate (Scan parameters) & Real (CT scans) \\
\bottomrule
\end{tabularx}
\caption{
While most existing benchmarks focus on natural image classification under semantic or synthetic shifts, CURE-OOD uniquely addresses acquisition-induced covariate shifts in clinical imaging and represents the first benchmark for OOD detection in risk modeling.}
\label{tab:benchmarks_compare}
\end{table*}

\subsection{Survival Analysis}
\label{sec:mtlr}

Survival prediction aims to estimate the likelihood of a clinical event, such as cancer recurrence or death, occurring over time. 
Unlike conventional classification tasks that output categorical labels, survival prediction models estimate a time-dependent probability distribution characterizing the risk of an event at each time point. 
This task is central to patient prognosis, as it enables individualized risk estimation beyond population-level measures such as five-year survival or median survival time. 
Accurate survival modeling provides clinicians with quantitative evidence to guide treatment selection and follow-up scheduling, making it a critical component of precision oncology. 
In cancer imaging, survival prediction relies on visual features extracted from medical scans to capture patterns associated with disease progression, supporting data-driven prognosis and treatment planning.

Traditional survival analysis methods, such as the Cox proportional hazards model, assume a fixed linear relationship between covariates and risk over time. 
This proportional hazards assumption limits their ability to capture non-proportional or time-varying effects commonly observed in real-world medical data. 
Moreover, these models are not designed to explicitly handle censored observations, where patients remain event-free during follow-up and the exact event time is unknown. 
To overcome these limitations, the multi-task logistic regression (MTLR) framework~\citep{yu2011learning} reformulates survival analysis as a series of dependent binary classification tasks across discretized time intervals. 
This formulation enables learning from both complete and censored samples within a unified probabilistic framework.

MTLR divides the continuous time axis into $m$ discrete intervals $\{t_1, t_2, \dots, t_m\}$ and learning conditional probabilities across them. 
Each patient’s survival status is encoded as a binary sequence $y = (y_1, y_2, \dots, y_m)$, where $y_i = 0$ indicates survival up to $t_i$ and $y_i = 1$ otherwise.
For each interval $t_i$, MTLR defines a logistic regression model:
\[
P_{\theta_i}(T \ge t_i \mid \mathbf{x}) = \left( 1 + \exp(\mathbf{\theta}_i^\top \mathbf{x} + b_i) \right)^{-1},
\]
where $\mathbf{x}$ denotes the imaging feature extracted by the visual backbone. 
All intervals are jointly optimized by enforcing temporal consistency, leading to the overall likelihood:
\begin{align}
P_{\Theta}(Y = y \mid \mathbf{x}) 
&= 
\frac{
\exp\!\left( \sum_{i=1}^{m} y_i(\mathbf{\theta}_i^\top \mathbf{x} + b_i) \right)
}{
\sum_{k=0}^{m} \exp(f_{\Theta}(\mathbf{x}, k))
},
\notag\\[2pt]
&\text{where } f_{\Theta}(\mathbf{x}, k) = \sum_{i=k+1}^{m} (\mathbf{\theta}_i^\top \mathbf{x} + b_i).
\end{align}
The resulting logits $f_{\Theta}(\mathbf{x}, k)$ represent the model’s estimated event likelihoods at different time intervals, providing a continuous and interpretable measure of risk progression over time.

\subsection{Medical Image Acquisition in Survival Analysis}
In computed tomography (CT), X-rays emitted from the tube pass through the patient and are attenuated by tissues of different densities before reaching the detector array. 
The detected signals are reconstructed into volumetric images in Hounsfield units. 
Scanner configuration and acquisition settings, including pixel spacing, exposure time, slice thickness, and X-ray tube current, determine key image properties such as spatial resolution, contrast, and noise. 
Variations in these parameters may result from hardware upgrades, calibration differences, or protocol adjustments across institutions, leading to observable changes in image texture and intensity profiles.
In survival prediction studies, datasets are often collected over long periods, sometimes spanning several years or even decades. 
During this time, imaging equipment and acquisition protocols are frequently updated, making such variability particularly common.
Details of these four parameters are further discussed in Sec.~\ref{sec:parameter}.

\subsection{Impact of Acquisition Variability}
Differences in acquisition parameters across scanners or institutions introduce systematic variability in resolution, contrast, and noise. 
These parameter-induced changes shift the underlying data distribution and create domain gaps between scans acquired under different conditions. 
In survival prediction, such covariate shifts can degrade model performance and lead to biased or unstable risk estimates. 
Understanding and quantifying the impact of these acquisition-induced shifts is therefore essential for developing robust and trustworthy prognostic models.

\else

\setlength{\abovedisplayskip}{3pt}
\setlength{\belowdisplayskip}{3pt}
\setlength{\abovedisplayshortskip}{2pt}
\setlength{\belowdisplayshortskip}{2pt}

\section{Background}
\label{sec:bg}

\begin{table}[t]
\centering
\eightpt
\setlength{\tabcolsep}{3pt}
    \caption{Comparison with existing OOD benchmarks. While most existing benchmarks focus on natural image classification under semantic or synthetic shifts, CURE-OOD addresses acquisition-induced covariate shifts in clinical imaging for survival prediction.}

\renewcommand{\arraystretch}{1}
\begin{tabularx}{\textwidth}{@{}l c >{\raggedright\arraybackslash}X >{\raggedright\arraybackslash}X l@{}}
\toprule
\textbf{Benchmark} & \textbf{Med.} & \textbf{Primary Task} & \textbf{Shift Type} & \textbf{Source} \\
\midrule
OpenOOD \citep{yang2022openood}& \xmark & Classification & Semantic & Mixed \\
OpenMIBOOD \citep{gutbrod2025openmibood}& \cmark & Classif. / Segmen. & Mixed & Real \\
ImageNet-ES \citep{baek2024unexplored}& \xmark & Classification & Covariate & Real \\
ImageNet-C/ CIFAR-C \citep{hendrycks2019benchmarking}& \xmark & Classification & Covariate & Synthetic \\
ImageNet-A/-O/-R \citep{hendrycks2021natural}& \xmark & Classification & Semantic & Real \\
WILDS \citep{pmlrv139koh21a} & Part. & Classif. / Regress. & Covariate & Real \\
MIDOG \citep{AUBREVILLE2023102699} & \cmark & Detection & Covariate & Real \\
PhaKIR \citep{rueckert2024phakir} & \cmark & Detect. / Classif. & Covariate & Real \\
OASIS-3 \citep{lamontagne2019oasis} & \cmark & Classification & Covariate & Real \\
MedMNIST v2 \citep{yang2023medmnist} & \cmark & Classification & Mixed & Mixed \\
BREEDS \citep{santurkar2020breeds} & \xmark & Classification & Subpopulation & Real \\
\midrule
\textbf{CURE-OOD} & \cmark & Risk modeling & Covariate & Real \\
\bottomrule
\end{tabularx}
\label{tab:benchmarks_compare}
\vspace{-10pt}
\end{table}

\subsection{Survival Analysis}
\label{sec:mtlr}

Survival prediction aims to estimate the likelihood of a clinical event, such as cancer recurrence or death, occurring over time.
Unlike standard classification tasks that predict discrete labels, survival models estimate a time-dependent probability distribution describing event risk across time intervals.
In medical imaging, such models leverage features extracted from CT scans to capture patterns associated with disease progression.
Among existing formulations, multi-task logistic regression (MTLR)~\citep{yu2011learning} is particularly suitable for censored survival prediction because it represents the task as a sequence of dependent binary classification tasks across discretized time intervals.

Let $\{t_1,\dots,t_m\}$ denote $m$ time intervals and $\mathbf{x}$ the features extracted by the visual backbone.
MTLR parameterizes the survival probability at each interval as
$P_{\theta_i}(T \ge t_i \mid \mathbf{x})=(1+\exp(\boldsymbol{\theta}_i^\top\mathbf{x}+b_i))^{-1}$,
and enforces temporal consistency through the joint likelihood
\begin{equation}
P_{\Theta}(Y=y\mid \mathbf{x})
=\frac{\exp\!\left(\sum_{i=1}^{m} y_i(\boldsymbol{\theta}_i^\top \mathbf{x}+b_i)\right)}
{\sum_{k=0}^{m}\exp\!\left(f_{\Theta}(\mathbf{x},k)\right)},
\quad
f_{\Theta}(\mathbf{x},k)=\sum_{i=k+1}^{m}(\boldsymbol{\theta}_i^\top \mathbf{x}+b_i).
\end{equation}
where the resulting logits $f_{\Theta}(\mathbf{x}, k)$ summarize interval-wise risk over time.
Model parameters are learned by minimizing the negative log-likelihood of this distribution.

From the predicted distribution, the event probability is
$p_{\Theta}(\mathbf{x},t_i)=P_{\Theta}(T=t_i\mid \mathbf{x})$,
the survival probability is
$G_{\Theta}(\mathbf{x},t_i)=P_{\Theta}(T\ge t_i\mid \mathbf{x})$,
and the discrete hazard is
\begin{equation}
h(t_i\mid \mathbf{x})
=\frac{P_{\Theta}(T=t_i\mid \mathbf{x})}{P_{\Theta}(T\ge t_i\mid \mathbf{x})}
=\frac{p_{\Theta}(\mathbf{x},t_i)}{G_{\Theta}(\mathbf{x},t_i)}.
\end{equation}

\subsection{Medical Image Acquisition for Survival Analysis}

CT appearance is determined by both acquisition and reconstruction settings.
Acquisition parameters directly affect image resolution, contrast, and noise characteristics. 
Variations in these settings, caused by scanner differences or protocol changes across sites and time, lead to systematic changes in image appearance, introducing covariate shifts. 
Therefore, we can leverage these parameters to define structured distribution shifts. 
We leverage this property to construct controlled ID and OOD splits, as detailed in Sec.~\ref{sec:parameter}.

\fi

\section{CURE-OOD Benchmark}

\subsection{Data and Preparation}
To construct CURE-OOD with controlled acquisition-induced shifts, we require a CT cohort that (i) provides scanner metadata for partitioning and (ii) supports standardized survival endpoints.
We therefore construct CURE-OOD by curating and partitioning the RADCURE cohort \citep{welch2024radcure}, which provides head and neck cancer (HNC) radiotherapy planning CT scans collected between 2005 and 2017.
Acquisition parameters are recorded in the DICOM metadata, including KVP, pixel spacing, exposure time, slice thickness, and X-ray tube current.
Since all scans were acquired with the same KVP setting, we exclude KVP from the partitioning design.
After filtering cases with incomplete CT volumes or missing primary GTVp masks, we retain 2,366 patients for benchmark construction.

Using this filtered cohort, we consider four survival prediction tasks: overall survival (OS), local failure-free survival (LFFS), regional failure-free survival (RFFS), and distant failure-free survival (DFFS).
The corresponding numbers of events are 792, 300, 134, and 293, respectively, and we treat each task independently during training and evaluation.

\begin{table}[t]
\centering
\eightpt
\setlength{\tabcolsep}{5pt}
\caption{Dataset partitioning based on acquisition parameters for CURE-OOD.}
\renewcommand{\arraystretch}{1}
\begin{tabular}{l c c c c c}
\toprule
\textbf{Parameter} & \textbf{Train} & \textbf{ID} & \textbf{OOD} & \textbf{ID Range} & \textbf{OOD Range} \\
\midrule
Pixel spacing (mm)  & 1293 & 100 & 100 & 0.976--1.172 & 0.805--0.969 \\
Exposure time (ms)  & 1288 & 100 & 100 & 1000 & 1831--2289 \\
Slice thickness (mm)& 1200 & 100 & 100 & 2 & 2.5 \\
X-ray tube current  & 1541 & 100 & 100 & 300 & 305--490 \\
\bottomrule
\end{tabular}
\label{tab:partition_rules}
\vspace{-10pt}
\end{table}

\subsection{Key Acquisition Parameters}
\label{sec:parameter}
We focus on four CT acquisition parameters that directly influence image appearance and feature distributions: pixel spacing, exposure time, slice thickness, and X-ray tube current. Together, they control effective resolution and noise, leading to realistic acquisition-induced shifts across sites and time (Fig.~\ref{fig:datashift}).

\textbf{Pixel Spacing} sets the spatial sampling resolution.
Smaller spacing preserves finer anatomical details but can amplify noise, while larger spacing smooths textures and yields a wider field of view.

\textbf{Exposure Time} controls photon accumulation at the detector.
Longer exposure generally improves signal-to-noise ratio and contrast, whereas shorter exposure increases noise and may obscure subtle structures.

\textbf{Slice Thickness} determines the degree of through-plane averaging.
Thicker slices reduce random noise but blur edges and fine structures, while thinner slices preserve sharper boundaries at the cost of higher noise.

\textbf{X-ray Tube Current} reflects the X-ray intensity.
Higher current typically produces cleaner, higher-contrast scans, whereas lower current leads to dimmer and noisier images.

These parameters are recorded in the RADCURE metadata and provide interpretable, reproducible axes for defining clinically realistic distribution shifts.

\subsection{Data Partitioning Strategy}
For each acquisition parameter, we first sort scans by the recorded value and group them according to distinct settings.
We then construct an ID/OOD split according to the natural distribution of the target acquisition parameter, assigning the larger group to the ID domain to ensure sufficient training data and the smaller group to the OOD domain.
For instance, for slice thickness, 2\,mm scans (1,614) form the ID set and 2.5\,mm scans (752) form the OOD set.

\begin{figure*}[t]
    \centering
    \begin{minipage}[t]{0.49\textwidth}
        \centering
        \begin{minipage}[c][0.23\textheight][c]{\linewidth}
            \centering
            \includegraphics[width=\linewidth,height=0.23\textheight,keepaspectratio]{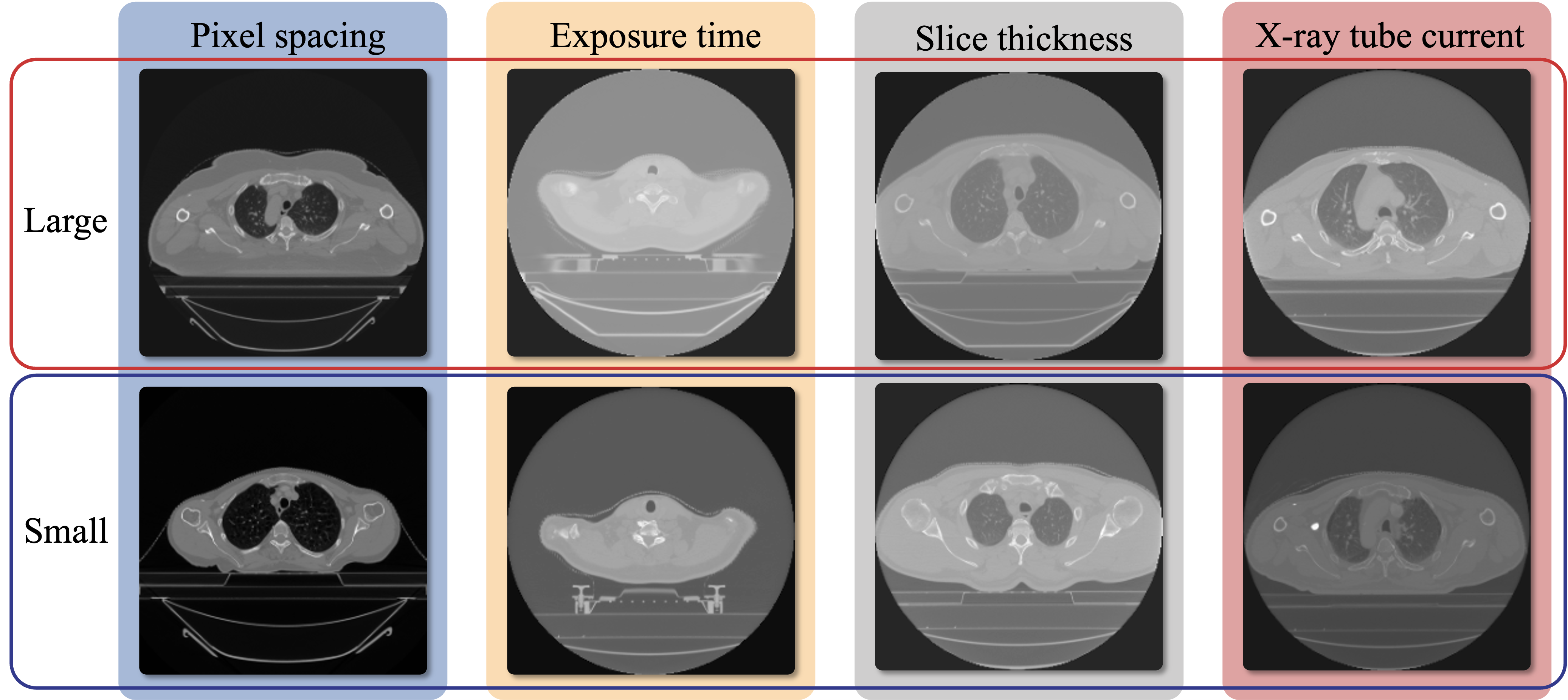}
        \end{minipage}
        \vspace{4pt}
        \captionof{figure}{Examples showing how acquisition parameters affect CT image appearance. Larger pixel spacing covers a wider field of view with decreased resolution, longer exposure time and higher tube current reduce noise, and thicker slice thickness produces smoother but blurrier images.}
        \label{fig:datashift}
    \end{minipage}\hfill
    \begin{minipage}[t]{0.49\textwidth}
        \centering
        \begin{minipage}[c][0.23\textheight][c]{\linewidth}
            \centering
            \includegraphics[width=\linewidth,height=0.23\textheight,keepaspectratio]{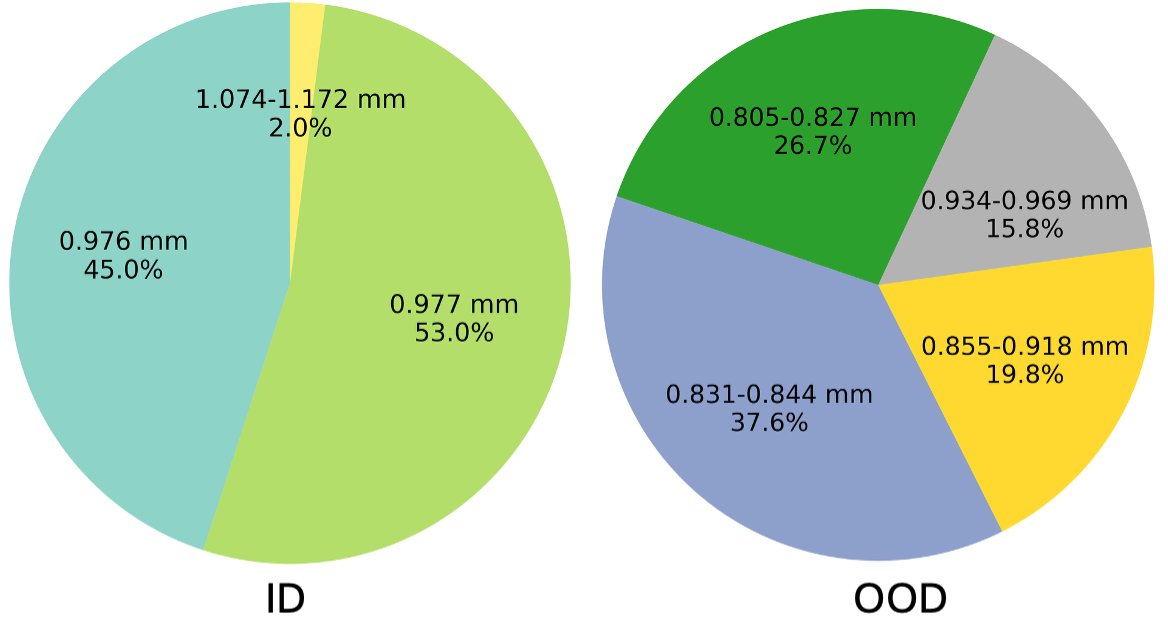}
        \end{minipage}
        \vspace{4pt}
        \captionof{figure}{Overall distribution of pixel spacing values across the RADCURE dataset. The variability across parameter ranges enables a clear distinction between ID and OOD domains used in CURE-OOD.}
        \label{fig:pixelspacing_distribution}
    \end{minipage}
\end{figure*}

Fig.~\ref{fig:pixelspacing_distribution} visualizes the resulting separation for pixel spacing.
Additional distributions for the other acquisition parameters are provided in Fig.~\ref{fig:datadistribution} in the appendix.
The ID/OOD assignment rules are summarized in Table~\ref{tab:partition_rules}.
For each task, samples without the corresponding acquisition metadata are excluded before partitioning.
We randomly sample 100 cases from each group to form the ID and OOD test sets.
The same partitioning strategy is applied to all four acquisition parameters, and the training set size ranges from approximately 1,200 to 1,600 cases across different splits.

\subsection{A Survival-Aware Reference Baseline}
%
Because MTLR does not produce class confidence scores, classification-oriented OOD scores are not naturally aligned with its outputs.
We therefore include \textbf{HazardDev} as a simple survival-aware reference baseline defined on a survival-native quantity.
We use hazard as the reference signal because it is directly derived from the predicted survival distribution and provides an interpretable summary of how event risk evolves over time. HazardDev then assigns an OOD score by measuring how much a sample's predicted hazard curve deviates from the average hazard curve on the training distribution.

\paragraph{Training-phase statistics.}
Given a trained MTLR model and the discrete-time hazard definition in Sec.~\ref{sec:mtlr}, we first compute the hazard function $h_n(t_i)$ for each training sample $n$ at each discretized time bin $t_i$.
We then calculate the average hazard over all $N$ training samples to obtain the expected hazard profile of the training distribution:
\begin{equation}
\bar{h}(t_i) = \frac{1}{N}\sum_{n=1}^{N} h_n(t_i).
\end{equation}

\paragraph{Testing-phase deviation.}
For a test sample with hazard prediction $h_{\text{test}}(t_i)$, we measure its deviation from the training distribution by computing the difference at each time bin and aggregating across all bins:
\begin{equation}
\text{HazardDev}(\mathbf{x}) = \sum_{i=1}^{m} h_{\text{test}}(t_i) - \bar{h}(t_i).
\end{equation}
A larger HazardDev value indicates that the sample's hazard pattern differs substantially from the training population, suggesting a higher likelihood of being out-of-distribution.

\ifnum\PaperVersion=1
    \section{Experiments}
This section presents our experimental protocol. We first introduce the evaluation metrics for survival prediction and OOD detection. We then describe the training procedure for survival prediction models. Finally, we detail the experimental setup, including architecture and optimization.

\subsection{Evaluation Metrics}
\label{sec:metrics}

We evaluate our model using metrics from both survival prediction and OOD detection perspectives. 
For survival prediction, we adopt the concordance index (C-index) as the primary metric to measure the consistency between the predicted risk scores and the actual survival times. 
A higher C-index indicates better concordance, meaning that patients with higher predicted risk are more likely to experience earlier events.

For OOD detection, we use three standard metrics, including the area under the receiver operating characteristic curve (AUROC), the area under the precision–recall curve (AUPR), and the false positive rate at 95\% true positive rate (FPR95).
AUROC and AUPR measure the overall discriminative ability of the OOD detector, while FPR95 reflects the false alarm rate when 95\% of ID samples are correctly identified.

All reported results in the following tables are based on these four metrics, with the C-index used for evaluating survival prediction accuracy, and AUROC, AUPR, and FPR95 used for assessing OOD detection performance.

\subsection{Survival Prediction Training}
To enable a comprehensive evaluation of OOD detection, we train separate survival prediction models for each combination of acquisition parameter and survival task. Specifically, we consider the four acquisition parameters introduced above and four survival outcomes, OS, LFFS, RFFS, and DFFS. For every parameter–task pair, we train an independent model using the corresponding ID training set under the MTLR framework~\citep{yu2011learning}. This results in sixteen models in total, each optimized on the ID domain of one specific acquisition parameter for one survival outcome. During evaluation, each model is tested on both its associated ID and OOD test sets, enabling systematic assessment of task-level OOD detection performance under controlled distribution shifts.This design reflects real deployment, where data from different acquisition conditions may appear at test time.
 \subsection{Experimental Setup}
We adopt a Vision Transformer-based architecture for survival prediction. The model employs a UNETR~\citep{hatamizadeh2022unetr} encoder with a ViT backbone to extract 3D volumetric features from CT scans. The extracted features are aggregated and passed through two fully connected layers, followed by an MTLR module that outputs survival probability distributions over discretized time intervals.

All CT volumes are preprocessed to a uniform spatial resolution and cropped around the primary tumor region. The survival timeline is discretized into $K=8$ intervals following the MTLR framework~\citep{yu2011learning}. The model is optimized using AdamW with a weight decay of $0.01$ and a batch size of 32. The learning rate is set to $1\times10^{-3}$ for OS and LFFS, $8\times10^{-4}$ for RFFS, and $3\times10^{-4}$ for DFFS. Training proceeds for up to 400 epochs with early stopping based on validation loss. Data augmentation includes random 3D rotation and spatial shifting. All experiments are implemented in PyTorch and conducted on NVIDIA RTX A5000 GPUs.

\section{Results on CURE-OOD}
In this section, we present the experimental results on the proposed CURE-OOD benchmark.
We first demonstrate how existing OOD detection methods originally designed for classification tasks can be seamlessly integrated into the survival prediction framework for a fair comparison under acquisition-induced shifts.
We then introduce our proposed HazardDev, specifically tailored for identifying distributional discrepancies in survival modeling.
Finally, we discuss why certain methods fail or even exhibit reversed trends under challenging shift conditions.

\subsection{Validation of Partition Effectiveness}
To verify that our parameter-based partitioning introduces meaningful distribution shifts, we trained a baseline survival prediction model on the training set and evaluated it on both the ID and OOD test sets for each acquisition parameter. 
Table~\ref{tab:ci_comparison} reports the C-index for all four survival prediction tasks. 
Across all acquisition parameters, model performance consistently decreases on the OOD test sets, confirming that our partitions effectively capture acquisition-induced variability. 
For the OS task, the C-index drops by 0.0397--0.0611 across different parameters, indicating relatively stable performance. 
This stability may be attributed to the more balanced event distribution in OS, which allows the model to learn consistent risk patterns. 
In contrast, the other three tasks show more pronounced performance drops, often exceeding 0.1. This can be attributed to their more imbalanced event distributions, which make model training less stable and lead to reduced generalization on OOD data.
These results demonstrate that variations in pixel spacing, exposure time, slice thickness, and tube current significantly affect survival prediction reliability, validating the effectiveness of our benchmark partitioning strategy.

\subsection{Applying Existing OOD Methods to Survival Modeling}

Since the MTLR framework reformulates survival prediction as a sequence of dependent classification tasks (Sec.~\ref{sec:mtlr}), 
the resulting logits $f_{\Theta}(\mathbf{x}, k)$ can be interpreted as prediction scores similar to class logits in conventional classifiers. 
This connection enables existing OOD detection methods, originally developed for classification, to be directly applied to survival prediction without architectural modifications. 
We adopt this formulation to provide a fair and consistent evaluation of traditional OOD approaches under acquisition-induced distribution shifts.
The experimental results of these methods across all shift types are summarized in Table~\ref{tab:all_shift_metrics}.

\subsection{Why Some Methods Failed}

We observe that several logits-based OOD detection methods, such as Energy and SCALE, behave in the opposite direction on the survival prediction task. 
In these cases, ID samples are often misclassified as OOD, while OOD samples are identified as ID. 
This phenomenon arises from the characteristics of survival modeling. 
MTLR is trained by minimizing the negative log-likelihood, which aligns the logits of ID samples closely with their true survival labels.
The monotonic nature of the survival function further encourages lower (more negative) logits at earlier time points to represent higher survival probabilities.

Formally, the unnormalized logit at interval $k$ is
\[
f_{\Theta}(\mathbf{x},k)=\sum_{r=k+1}^{m}(\boldsymbol{\theta}_r^\top\mathbf{x}+b_r),
\]
and the per-sample gradient takes the form
\[
\frac{\partial \ell_i}{\partial \boldsymbol{\theta}_j}
=\Big(y_j(s_i)-\sum_{k<j}\pi_{ik}\Big)\mathbf{x}_i,
\qquad
\pi_{ik}=\frac{\exp f_{\Theta}(\mathbf{x}_i,k)}{\sum_{r}\exp f_{\Theta}(\mathbf{x}_i,r)}.
\]
At early intervals, most samples survive ($y_j\!\approx\!1$) while the model still assigns nonzero mass $\sum_{k<j}\pi_{ik}>0$, leading to $\Delta f_{\Theta}(\mathbf{x},k)\!<\!0$ for $k<j$. 
Hence, optimization naturally drives early logits toward smaller (more negative) values for in-distribution data.

When a distribution shift occurs, OOD samples lie in unseen regions of the feature space, reducing the projection magnitude $|\mathbf{\theta}_r^\top\mathbf{x}|$ and making the logits drift toward zero. 
As a result, ID logits become smaller (more negative) than OOD logits, which is the reverse of the conventional pattern in classification where ID logits are larger. 
This inversion leads logit-based scores such as Energy or SCALE to misidentify ID samples as OOD. 
The effect is empirically confirmed in Fig.~\ref{fig:os_logits_distribution} and further analyzed in Sec.~\ref{sec:reverse}.

\subsection{Hazard-Based OOD Detection}

We observe that the hazard values predicted by the survival model reveal clear distributional differences between ID and OOD samples. 
As shown in Fig.~\ref{fig:hazard_curves}, ID samples consistently exhibit higher hazard levels than OOD samples, indicating a clear separation between the two distributions.
Motivated by this observation, we introduce a simple method, HazardDev, which quantifies the deviation of a sample’s hazard curve from the expected training distribution. 
A larger deviation indicates a higher likelihood of being OOD. 
Across all four tasks, HazardDev achieves the best average OOD detection performance under different acquisition shifts, demonstrating its effectiveness and robustness.
Further formulation details and implementation are provided in the supplementary material.
\else
    \section{Experiments}

\parhead{Evaluation metrics.}
\label{sec:metrics}
We report metrics for both survival prediction and OOD detection. For survival prediction, we use the concordance index (C-index), where higher values indicate better agreement between predicted risk and observed time-to-event. 
For OOD detection, we use AUROC and AUPRC, measuring overall separability.

\parhead{Survival prediction training.}
To evaluate OOD detection under different distribution shifts, we train separate MTLR~\citep{yu2011learning} survival models for each acquisition parameter--outcome pair. We consider four acquisition parameters and four outcomes (OS, LFFS, RFFS, DFFS), resulting in 16 models. Each model is trained on its corresponding ID training set and evaluated on both the matched ID test set and the paired OOD test set, reflecting deployment where acquisition conditions may change at inference time.

\begin{table}[t]
\centering
\fontsize{6.6}{6.8}\selectfont
\setlength{\tabcolsep}{0.8pt}
\renewcommand{\arraystretch}{1}
\caption{\footnotesize OOD detection performance across survival tasks measured by AUROC and AUPRC. Average results underscore HazardDev's robustness under acquisition shifts.}
\begin{tabularx}{\textwidth}{l@{\hspace{3pt}}l *{10}{>{\centering\arraybackslash}X}}
\toprule
\multirow{2}{*}{} & \multirow{2}{*}{\textbf{Method}} &
\multicolumn{2}{c}{Pixel Spacing} & \multicolumn{2}{c}{Exposure Time} & \multicolumn{2}{c}{Slice Thick.} & \multicolumn{2}{c}{Tube Current} & \multicolumn{2}{c}{\textbf{Average}} \\
\cmidrule(lr){3-4} \cmidrule(lr){5-6} \cmidrule(lr){7-8} \cmidrule(lr){9-10} \cmidrule(lr){11-12}
& & AUROC & AUPRC & AUROC & AUPRC & AUROC & AUPRC & AUROC & AUPRC & AUROC & AUPRC \\
\midrule
\multirow{11}{*}{\rotatebox{90}{OS}}
& ASH                & 0.3910 & 0.4505 & 0.3815 & 0.4354 & 0.4670 & 0.4783 & 0.4381 & 0.4795 & 0.4194 & 0.4609 \\
& Dropout            & 0.5353 & 0.5040 & 0.4040 & 0.4458 & 0.4778 & 0.4847 & 0.4711 & 0.4851 & 0.4721 & 0.4799 \\
& Energy             & 0.3938 & 0.4534 & 0.3807 & 0.4300 & 0.4594 & 0.4779 & 0.4201 & 0.4745 & 0.4135 & 0.4589 \\
& GEN                & 0.5411 & 0.5101 & 0.4223 & 0.4441 & 0.4843 & 0.4909 & 0.5074 & \underline{0.5105} & 0.4888 & 0.4889 \\
& Dice               & 0.4200 & 0.4631 & 0.4001 & 0.4370 & 0.4615 & 0.4789 & 0.4329 & 0.4763 & 0.4286 & 0.4638 \\
& MLS                & 0.4073 & 0.4586 & 0.3980 & 0.4366 & 0.4580 & 0.4780 & 0.4207 & 0.4728 & 0.4210 & 0.4615 \\
& MSP                & \underline{0.5472} & \underline{0.5121} & \underline{0.4279} & \underline{0.4466} & \underline{0.4995} & \underline{0.4983} & \underline{0.5120} & 0.5098 & \underline{0.4967} & \underline{0.4917} \\
& ODIN               & 0.5169 & 0.4980 & 0.4070 & 0.4320 & 0.4784 & 0.4819 & 0.4912 & 0.5043 & 0.4734 & 0.4790 \\
& SCALE              & 0.3994 & 0.4544 & 0.3842 & 0.4384 & 0.4635 & 0.4801 & 0.4294 & 0.4798 & 0.4191 & 0.4632 \\
& \textbf{HazardDev} & \textbf{0.6043} & \textbf{0.5726} & \textbf{0.6237} & \textbf{0.6098} & \textbf{0.5416} & \textbf{0.5455} & \textbf{0.5797} & \textbf{0.5465} & \textbf{0.5873} & \textbf{0.5686} \\
\midrule
\multirow{11}{*}{\rotatebox{90}{LFFS}}
& ASH                & 0.4056 & 0.4480 & 0.4284 & 0.4736 & 0.4458 & 0.4580 & 0.4686 & 0.5104 & 0.4371 & 0.4725 \\
& Dropout            & 0.3931 & 0.4378 & 0.4433 & 0.4932 & 0.4566 & 0.4659 & 0.5014 & 0.5226 & 0.4486 & 0.4799 \\
& Energy             & 0.3942 & 0.4400 & 0.3578 & 0.4028 & \underline{0.4661} & 0.4658 & 0.4521 & 0.5044 & 0.4175 & 0.4532 \\
& GEN                & \underline{0.4551} & \underline{0.4634} & \underline{0.6333} & \underline{0.5950} & 0.4490 & \underline{0.4837} & \underline{0.5483} & \textbf{0.5506} & \underline{0.5214} & \underline{0.5232} \\
& Dice               & 0.3991 & 0.4424 & 0.4579 & 0.4951 & 0.4461 & 0.4626 & 0.4575 & 0.5081 & 0.4401 & 0.4771 \\
& MLS                & 0.3953 & 0.4406 & 0.4319 & 0.4790 & 0.4607 & 0.4684 & 0.4642 & 0.5115 & 0.4380 & 0.4749 \\
& MSP                & 0.4376 & 0.4550 & 0.6242 & 0.5891 & 0.4447 & 0.4688 & 0.5192 & \underline{0.5348} & 0.5064 & 0.5119 \\
& ODIN               & 0.4397 & 0.4548 & 0.6097 & 0.5708 & 0.4638 & 0.4736 & 0.4754 & 0.5100 & 0.4971 & 0.5023 \\
& SCALE              & 0.3966 & 0.4410 & 0.4160 & 0.4655 & 0.4607 & 0.4679 & 0.4775 & 0.5159 & 0.4377 & 0.4726 \\
& \textbf{HazardDev} & \textbf{0.6082} & \textbf{0.5951} & \textbf{0.6459} & \textbf{0.6607} & \textbf{0.5347} & \textbf{0.5864} & \textbf{0.5527} & 0.5203 & \textbf{0.5854} & \textbf{0.5906} \\
\midrule
\multirow{11}{*}{\rotatebox{90}{RFFS}}
& ASH                & 0.4653 & 0.4834 & 0.4197 & 0.4672 & 0.4850 & 0.4836 & 0.5410 & 0.5405 & 0.4778 & 0.4937 \\
& Dropout            & 0.4310 & 0.4474 & 0.3781 & 0.4161 & 0.5147 & 0.5397 & 0.5408 & 0.5296 & 0.4661 & 0.4832 \\
& Energy             & 0.4128 & 0.4372 & 0.4019 & 0.4219 & 0.4879 & 0.5073 & 0.5088 & 0.5018 & 0.4528 & 0.4671 \\
& GEN                & \underline{0.5438} & \underline{0.5143} & 0.5305 & 0.5208 & 0.5272 & \textbf{0.5757} & \underline{0.5417} & \underline{0.5660} & \underline{0.5358} & \underline{0.5442} \\
& Dice               & 0.4317 & 0.4457 & 0.3930 & 0.4203 & \textbf{0.5288} & 0.5359 & 0.5313 & 0.5186 & 0.4712 & 0.4801 \\
& MLS                & 0.4271 & 0.4446 & 0.3972 & 0.4215 & \textbf{0.5288} & 0.5395 & 0.5366 & 0.5242 & 0.4724 & 0.4824 \\
& MSP                & 0.4998 & 0.4848 & \underline{0.5517} & \underline{0.5216} & 0.5084 & \underline{0.5651} & 0.5378 & \textbf{0.5671} & 0.5244 & 0.5346 \\
& ODIN               & 0.5066 & 0.4927 & 0.5085 & 0.5026 & 0.4950 & 0.5520 & 0.5194 & 0.5471 & 0.5074 & 0.5236 \\
& SCALE              & 0.4330 & 0.4558 & 0.3978 & 0.4198 & 0.5074 & 0.5169 & 0.5398 & 0.5354 & 0.4695 & 0.4820 \\
& \textbf{HazardDev} & \textbf{0.6060} & \textbf{0.6089} & \textbf{0.5984} & \textbf{0.6421} & \underline{0.5233} & 0.5112 & \textbf{0.5500} & 0.5376 & \textbf{0.5694} & \textbf{0.5749} \\
\midrule
\multirow{11}{*}{\rotatebox{90}{DFFS}}
& ASH                & 0.4608 & 0.4653 & 0.3638 & 0.4125 & 0.4811 & 0.5121 & 0.5170 & 0.5179 & 0.4557 & 0.4769 \\
& Dropout            & 0.5387 & 0.5197 & 0.4728 & 0.4528 & \textbf{0.5570} & \underline{0.5449} & 0.4918 & 0.5086 & 0.5151 & 0.5065 \\
& Energy             & 0.4633 & 0.4713 & 0.3579 & 0.4067 & 0.4699 & 0.5086 & 0.4878 & 0.5066 & 0.4447 & 0.4733 \\
& GEN                & 0.5420 & 0.5168 & \underline{0.5543} & \underline{0.4964} & 0.5486 & 0.5437 & 0.5374 & \underline{0.5235} & 0.5456 & 0.5201 \\
& Dice               & 0.5316 & 0.5112 & 0.4162 & 0.4282 & 0.5376 & 0.5377 & 0.4911 & 0.5074 & 0.4941 & 0.4961 \\
& MLS                & 0.4725 & 0.4787 & 0.3861 & 0.4181 & 0.4656 & 0.5070 & 0.5046 & 0.5107 & 0.4572 & 0.4786 \\
& MSP                & \textbf{0.5587} & 0.5277 & 0.5314 & 0.4800 & \underline{0.5522} & \textbf{0.5483} & \underline{0.5483} & \textbf{0.5258} & \underline{0.5476} & \underline{0.5204} \\
& ODIN               & \underline{0.5531} & \underline{0.5380} & 0.4945 & 0.4605 & 0.5074 & 0.5173 & \textbf{0.5514} & 0.5171 & 0.5266 & 0.5082 \\
& SCALE              & 0.4668 & 0.4738 & 0.3606 & 0.4108 & 0.4832 & 0.5114 & 0.5166 & 0.5141 & 0.4568 & 0.4775 \\
& \textbf{HazardDev} & 0.5386 & \textbf{0.5554} & \textbf{0.6464} & \textbf{0.6630} & 0.5338 & 0.5134 & 0.5201 & 0.5152 & \textbf{0.5597} & \textbf{0.5617} \\
\bottomrule
\end{tabularx}
\label{tab:all_shift_metrics}
\vspace{-10pt}
\end{table}

\parhead{Experimental setup.}
We use a ViT-based 3D model with a UNETR~\citep{hatamizadeh2022unetr} encoder to extract volumetric features from CT, followed by two fully connected layers and an MTLR head that outputs survival distributions over discretized time. CT volumes are resampled to a common resolution and cropped around the primary tumor; the timeline is discretized into 8 intervals. We train with AdamW (weight decay $0.01$, batch size 32) for up to 400 epochs with early stopping on validation loss; learning rates are $1{\times}10^{-3}$ (OS/LFFS), $8{\times}10^{-4}$ (RFFS), and $3{\times}10^{-4}$ (DFFS). Augmentation includes random 3D rotation and shifts. All experiments use PyTorch on NVIDIA RTX A5000 GPUs.

\section{Results on CURE-OOD}
\subsection{Performance Degradation under Data Shifts}
To verify the impact of acquisition-induced shifts on downstream survival performance, we compare C-index on the ID and OOD test sets for each acquisition parameter. Table~\ref{tab:ci_comparison} reports C-index results across four tasks. 
When the ID-trained model is evaluated on OOD test data, survival accuracy generally decreases, showing that acquisition changes can degrade predictive performance. 
For OS, the degradation is modest for three shifts (0.0397--0.0611) but nearly disappears under slice thickness (0.0013), suggesting that the effect of acquisition shift is task- and parameter-dependent. In contrast, LFFS, RFFS, and DFFS show larger drops, often $>0.1$, which may reflect the smaller number of events and the resulting class imbalance, making the models less robust under shifts. 
Overall, these results suggest that robustness under acquisition shift depends not only on the shift itself but also on endpoint-specific event characteristics.
Together, the observed degradation patterns suggest that distribution shifts can have measurable downstream consequences for survival prediction.

\begin{table}[t]
\centering
\eightpt
\caption{\footnotesize Comparison of C-index between ID and OOD test sets across four survival prediction tasks. The consistent performance decrease (Dec.) on OOD data underscores the impact of acquisition-induced distribution shifts on survival prediction accuracy.}
\setlength{\tabcolsep}{1.2pt}
\renewcommand{\arraystretch}{0.9}
\begin{tabularx}{\textwidth}{l *{12}{>{\centering\arraybackslash}X}}
\toprule
\multirow{2}{*}{\textbf{Task}} &
\multicolumn{3}{c}{\textbf{Pixel Spacing}} &
\multicolumn{3}{c}{\textbf{Exposure Time}} &
\multicolumn{3}{c}{\textbf{Slice Thickness}} &
\multicolumn{3}{c}{\textbf{Tube Current}} \\
\cmidrule(lr){2-4} \cmidrule(lr){5-7} \cmidrule(lr){8-10} \cmidrule(lr){11-13}
 & ID & OOD & Dec. & ID & OOD & Dec. & ID & OOD & Dec. & ID & OOD & Dec. \\
\midrule
OS   & 0.7163 & 0.6675 & 0.0488 & 0.7254 & 0.6857 & 0.0397 & 0.6583 & 0.6570 & 0.0013 & 0.7430 & 0.6819 & 0.0611 \\
LFFS & 0.7227 & 0.5969 & 0.1258 & 0.6431 & 0.5969 & 0.0462 & 0.7443 & 0.6937 & 0.0506 & 0.7411 & 0.6045 & 0.1366 \\
RFFS & 0.7974 & 0.7704 & 0.0270 & 0.8145 & 0.6125 & 0.2020 & 0.7405 & 0.5772 & 0.1633 & 0.6901 & 0.6376 & 0.0525 \\
DFFS & 0.7103 & 0.6784 & 0.0319 & 0.8912 & 0.5925 & 0.2987 & 0.7208 & 0.6751 & 0.0457 & 0.7646 & 0.6526 & 0.1120 \\
\bottomrule
\end{tabularx}
\label{tab:ci_comparison}
\vspace{-16pt}
\end{table}

\subsection{Why Classification-Style OOD Scores Fail}
Classification-style OOD detectors are typically designed for class-confidence scores, whereas survival prediction produces structured, time-dependent outputs. In this setting, these detectors can still be directly applied to interval-wise logits $f_{\Theta}(\mathbf{x}, k)$, enabling a fair comparison between conventional post-hoc OOD scores and survival-aware alternatives on the same benchmark. In practice, Table~\ref{tab:all_shift_metrics} shows that several mainstream logit-based methods, including Energy, ASH, and SCALE, do not provide reliable OOD signals in survival prediction; they often show weak discrimination and, in some settings, even reversed ordering, where ID samples receive more OOD-like scores than OOD samples.

The reason is that survival logits do not encode class confidence in the same way as classification logits. Instead, as described in Sec.~\ref{sec:mtlr}, they parameterize a time-ordered survival distribution and therefore reflect a structured, time-dependent risk profile. At early intervals, most patients have not yet experienced an event, so the training objective places greater emphasis on survival through those intervals, which in turn encourages the corresponding logits to move toward smaller, more negative values. Many classification-style OOD scores implicitly assume that larger logit magnitude or score values correspond to stronger in-distribution confidence. In survival prediction, however, larger logits are not necessarily more confident predictions, so the resulting score ordering need not follow the confidence assumptions used by classification-style OOD detectors.


\begin{figure}[t]
  \centering
  \begin{minipage}{1\linewidth}
    \centering
    \begin{minipage}[b]{0.49\linewidth}
      \includegraphics[width=\linewidth]{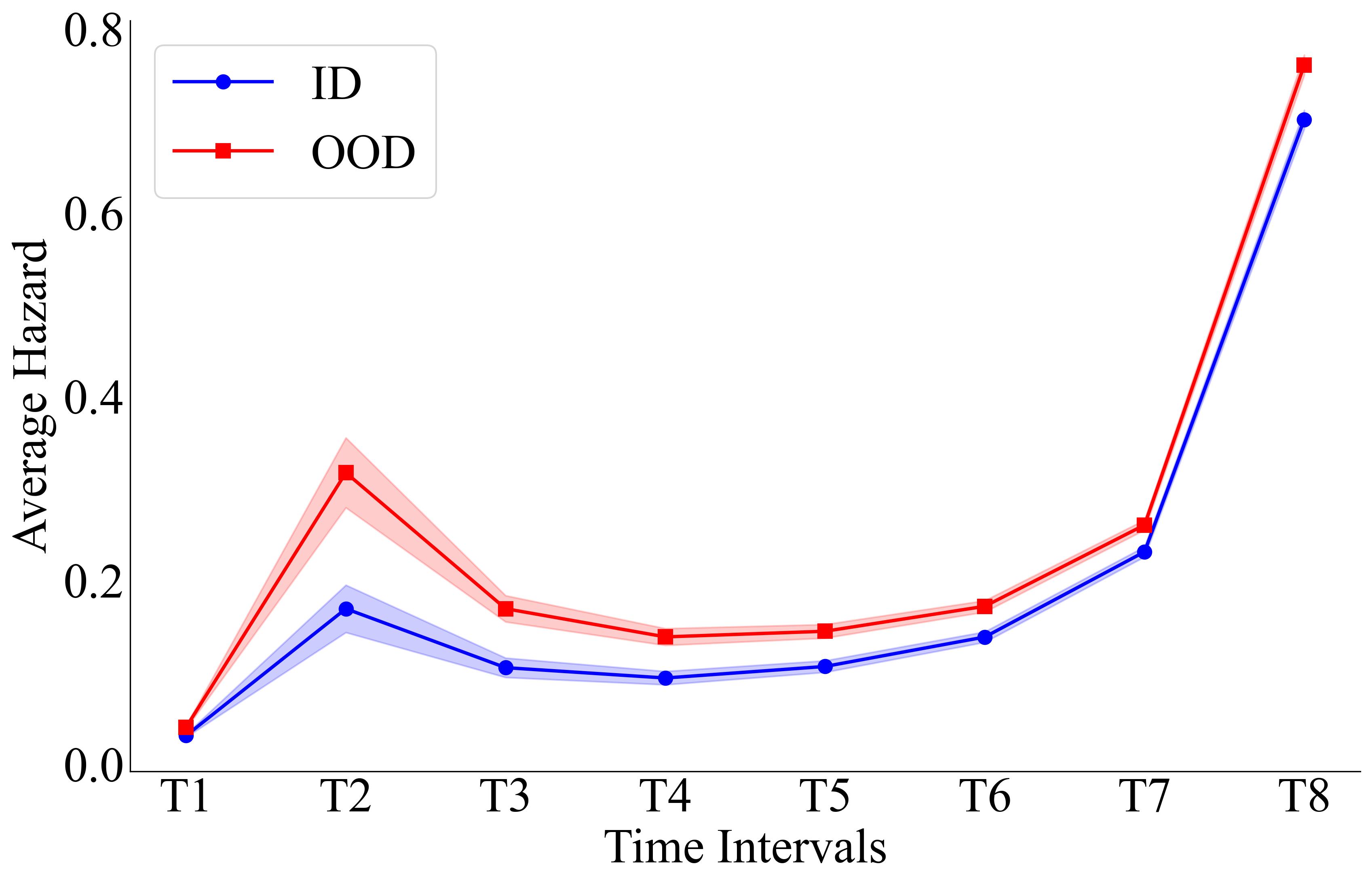}
      \vspace{2pt}

      \centering
      \footnotesize Exposure Time
    \end{minipage}\hfill
    \begin{minipage}[b]{0.49\linewidth}
      \centering
      \includegraphics[width=\linewidth]{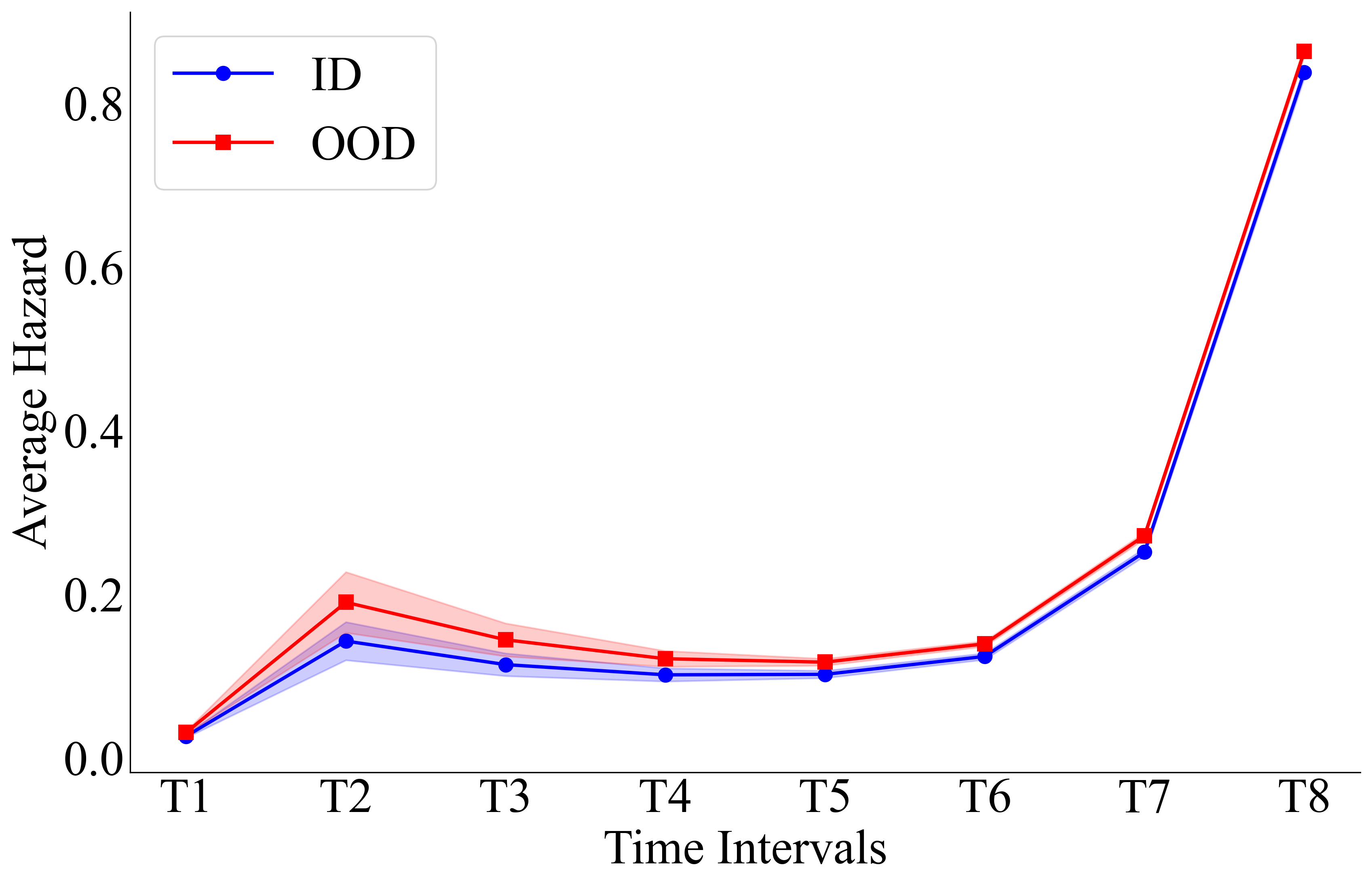}
      \vspace{2pt}

      \footnotesize Pixel Spacing
    \end{minipage}
  \end{minipage}
  \caption{\footnotesize Mean hazard curves for ID and OOD test sets on OS under two acquisition shifts. Shaded regions indicate $\pm$1 std. OOD shows higher mean hazard than ID, suggesting hazard magnitude is an OOD signal.}
  \label{fig:hazard_curves_2}
\end{figure}

Under acquisition-induced shift, OOD samples fall into less familiar feature regions, making the model more likely to assign higher event risk than it does for well-aligned ID samples. Their logits therefore become larger and drift toward zero. Consequently, ID samples often produce more negative logits than OOD samples, which is the opposite of the usual classification pattern where ID examples yield stronger, larger-magnitude evidence for the predicted class. When classification-style scores are applied on top of these logits, the ordering can invert: scores that are supposed to be larger for OOD inputs may instead become larger for ID inputs, causing systematic ID-as-OOD errors.

This effect is shown in Fig.~\ref{fig:mean_mtlr_logits}, where the average OOD logits are consistently higher and closer to zero than the corresponding ID logits under a representative acquisition shift. This is fundamentally different from standard classification, where ID samples typically induce stronger class evidence and more extreme logits than OOD samples. The logits distributions in Fig.~\ref{fig:os_logits_distribution} and the gradient analysis in Sec.~\ref{sec:reverse} further support the same mechanism. Together, these results explain why the evaluated classification-oriented OOD scores are not reliably transferable to survival prediction, even when the same methods are effective in standard classification benchmarks.

\subsection{Hazard-Based OOD Detection}
In contrast, HazardDev remains aligned with the output semantics of survival prediction by operating on the predicted hazard curve instead of raw logits. As shown in Table~\ref{tab:all_shift_metrics}, it provides the strongest average survival-aware reference among the compared methods across survival outcomes, with average AUROC improvements of 0.109, 0.062, 0.029, and 0.004 for OS, LFFS, RFFS, and DFFS, respectively, and average AUPR improvements of 0.095, 0.055, 0.022, and 0.004 in the same order. These results suggest that hazard-based deviation is a useful survival-aligned OOD signal under survival formulations.

To further probe this behavior, we visualize the mean hazard curves on ID vs.\ OOD test sets for OS under pixel spacing and exposure time shifts (Fig.~\ref{fig:hazard_curves_2}). Across all discretized intervals, OOD samples exhibit systematically higher mean hazard than ID samples, supporting the use of hazard-based deviation as a task-consistent OOD signal. More broadly, this suggests that future OOD methods for survival prediction should be defined on survival-native predictive quantities rather than directly transferred classification confidence scores.

\subsection{Architecture Sensitivity Analysis}
To explore HazardDev's performance with a different feature extractor, we use a ResNet model and repeat training and OOD detection on the OS task under the same four acquisition shifts. Table~\ref{tab:resnet_os_metrics} shows that the overall ranking changes with the feature extractor, which is expected because different backbones induce different feature geometries for post-hoc OOD scoring. However, the main qualitative trend remains: hazard-based scores stay competitive under the backbone change and continue to provide strong detection performance under acquisition shifts.

More specifically, the HazardDev score remains particularly effective on exposure time and slice thickness. Among the methods retained from Table~\ref{tab:all_shift_metrics}, it attains the highest AUROC and AUPRC on exposure time, the highest AUROC and AUPRC on slice thickness, and the strongest average AUROC/AUPRC at 0.5892/0.5804. These results indicate that the utility of hazard-based deviation is not tied to a single backbone design and that this survival-aware signal remains informative under a different feature extractor. In turn, this suggests that the main benchmark findings extend across different feature extractors and capture a broader property of survival OOD detection.

\begin{table}[t]
\centering
\fontsize{6.2}{6.4}\selectfont
\setlength{\tabcolsep}{0.55pt}
\renewcommand{\arraystretch}{1}
\caption{\footnotesize OOD detection results on the OS task with a ResNet backbone. We report AUROC and AUPRC across four acquisition shifts.}

\begin{tabularx}{\textwidth}{l@{\hspace{3pt}}l *{10}{>{\centering\arraybackslash}X}}
\toprule
\multirow{2}{*}{} & \multirow{2}{*}{\textbf{Method}} &
\multicolumn{2}{c}{Pixel Spacing} & \multicolumn{2}{c}{Exposure Time} & \multicolumn{2}{c}{Slice Thick.} & \multicolumn{2}{c}{Tube Current} & \multicolumn{2}{c}{\textbf{Average}} \\
\cmidrule(lr){3-4} \cmidrule(lr){5-6} \cmidrule(lr){7-8} \cmidrule(lr){9-10} \cmidrule(lr){11-12}
& & AUROC & AUPRC & AUROC & AUPRC & AUROC & AUPRC & AUROC & AUPRC & AUROC & AUPRC \\
\midrule
\multirow{10}{*}{\rotatebox{90}{OS}}
& ASH                & 0.4740 & 0.5079 & 0.3740 & 0.4167 & 0.4049 & 0.4357 & 0.4724 & 0.5100 & 0.4313 & 0.4676 \\
& Dropout            & 0.5101 & \textbf{0.5245} & 0.4111 & 0.4335 & 0.4592 & 0.4554 & 0.5457 & 0.5201 & 0.4815 & 0.4834 \\
& GEN                & 0.4965 & 0.5204 & 0.4708 & 0.4550 & 0.4828 & 0.4692 & 0.5405 & 0.5506 & 0.4976 & 0.4988 \\
& Energy             & 0.4782 & 0.4934 & 0.3440 & 0.4175 & 0.3865 & 0.4386 & 0.4443 & 0.4454 & 0.4133 & 0.4487 \\
& Dice               & 0.4673 & 0.5056 & 0.3853 & 0.4217 & 0.4139 & 0.4390 & 0.5565 & 0.5425 & 0.4557 & 0.4772 \\
& MLS                & 0.4782 & 0.5096 & 0.3439 & 0.4076 & 0.3861 & 0.4303 & 0.4447 & 0.5042 & 0.4132 & 0.4629 \\
& MSP                & 0.4862 & 0.5152 & 0.4985 & 0.4689 & 0.4956 & 0.4773 & 0.5392 & 0.5445 & 0.5049 & 0.5015 \\
& ODIN               & 0.4769 & 0.5110 & 0.4851 & 0.4616 & 0.4897 & 0.4715 & 0.5382 & 0.5214 & 0.4975 & 0.4914 \\
& SCALE              & 0.4771 & 0.5088 & 0.3604 & 0.4125 & 0.3797 & 0.4279 & 0.4754 & 0.5137 & 0.4232 & 0.4657 \\
& \textbf{HazardDev} & \textbf{0.5218} & 0.5118 & \textbf{0.6559} & \textbf{0.6446} & \textbf{0.6135} & \textbf{0.6140} & \textbf{0.5657} & \textbf{0.5510} & \textbf{0.5892} & \textbf{0.5804} \\
\bottomrule
\end{tabularx}
\label{tab:resnet_os_metrics}
\vspace{-8pt}
\end{table}

\begin{figure*}[t]
  \centering
  \begin{minipage}[t]{0.49\textwidth}
    \centering
    \includegraphics[width=\linewidth]{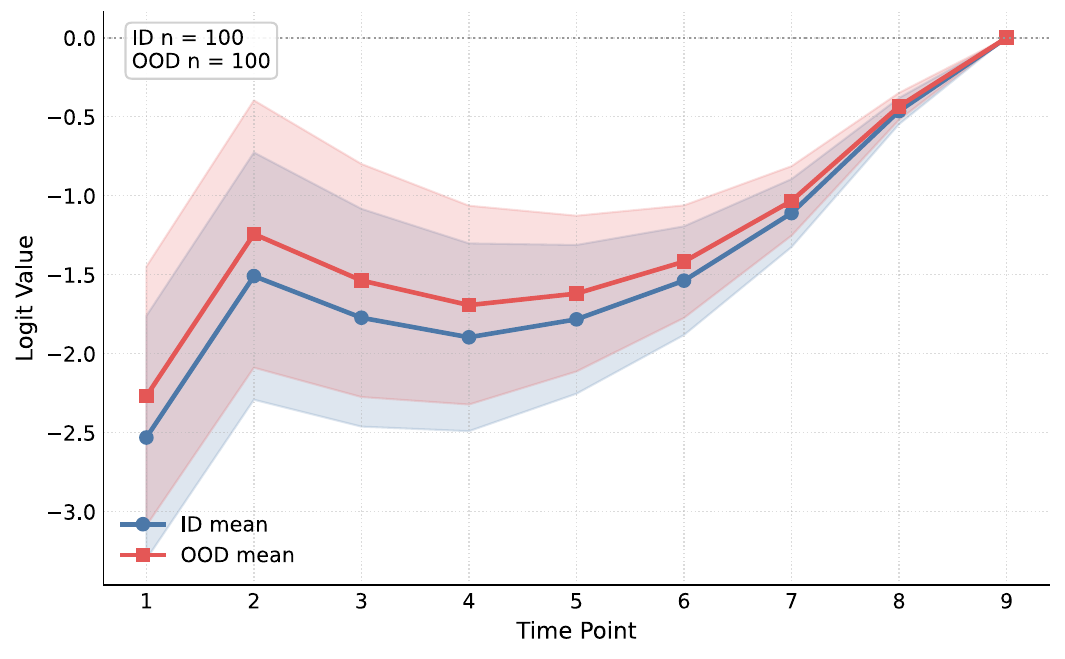}
    \captionof{figure}{\footnotesize Mean logits on the OS task under the exposure time shift. Unlike standard classification logits, ID samples produce systematically smaller, more negative logits across time intervals, while OOD samples drift toward zero. This reversed ordering breaks the confidence interpretation assumed by classification-style OOD scores.}
    \label{fig:mean_mtlr_logits}
  \end{minipage}\hfill
  \begin{minipage}[t]{0.49\textwidth}
    \centering
    \includegraphics[width=\linewidth]{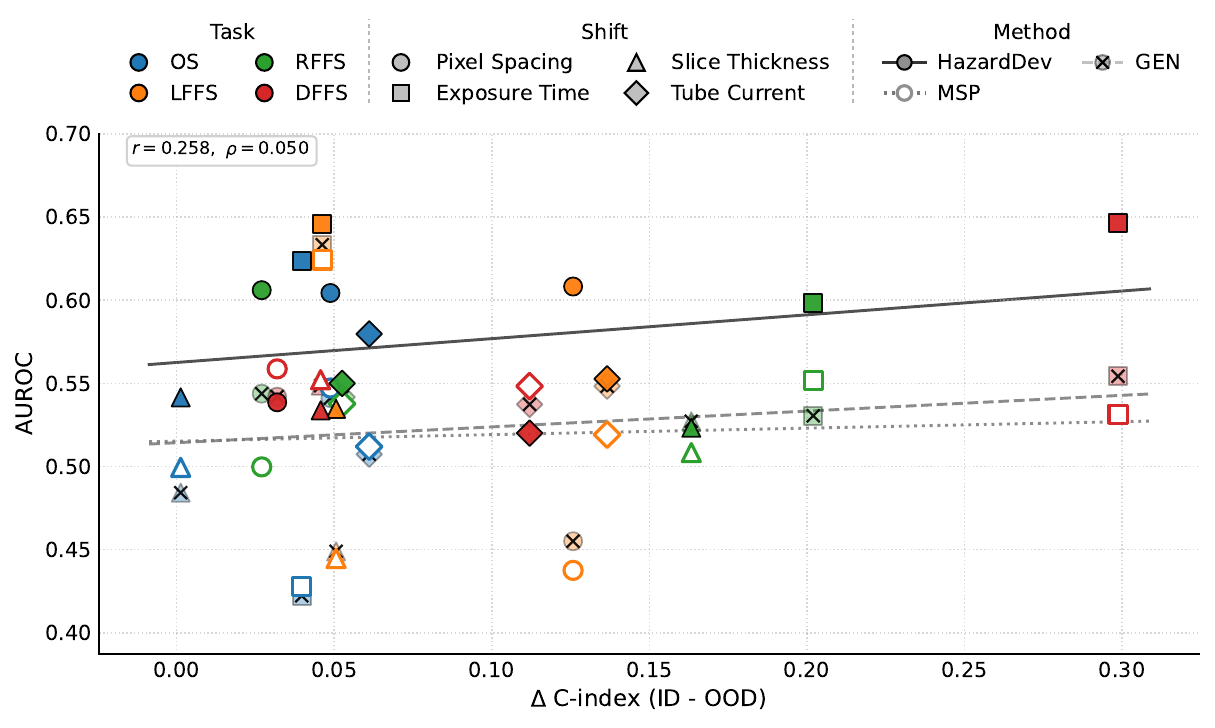}
    \captionof{figure}{\footnotesize Relationship between downstream survival degradation and OOD detection quality across task--shift pairs. Each point is one method on one task--shift combination. The reported $r$ and $\rho$ are Pearson and Spearman correlations for HazardDev, indicating weak linear and rank associations between OOD detectability and downstream performance loss.}
    \label{fig:drop_vs_auroc}
  \end{minipage}
\end{figure*}

\subsection{Limits of OOD Detection for Robust Survival Prediction}
We found that OOD separability does not necessarily reflect how much a shift harms the downstream survival task. To explore this relationship, we compare the C-index drop between ID and OOD test sets against OOD detection performance across all 16 task--shift combinations for three representative methods including HazardDev, GEN and MSP in Fig.~\ref{fig:drop_vs_auroc}. If stronger OOD signals always corresponded to larger deployment risk, points with larger $\Delta$ C-index would also show consistently higher OOD detection performance. Instead, the relationship is weak and unstable.

Across methods three methods,  we found that larger C-index drops do not reliably correspond to better OOD detection. Even for HazardDev, the Pearson and Spearman correlations are only 0.258 and 0.050, respectively. This further shows that larger C-index drops do not reliably correspond to higher OOD detection performance. Some shifts are easy to distinguish from ID data yet only modestly reduce survival accuracy, whereas others substantially degrade prediction while remaining only moderately separable. 
Taken together, these findings suggest that current methods remain limited in supporting robust survival prediction under acquisition-induced shifts.

\fi

\ifnum\PaperVersion=1
    \section{Conclusion and Future Work}
In this work, we introduced CURE-OOD, the first benchmark dedicated to OOD detection in cancer survival prediction. CURE-OOD systematically partitions CT scans by four scans parameters—pixel spacing, exposure time, slice thickness, and X-ray tube current—to simulate realistic covariate shifts. Through comprehensive experiments, we demonstrated that such acquisition-induced shifts substantially degrade survival prediction performance. We further evaluated a wide range of OOD detection methods originally designed for classification tasks and found that while some methods transfer reasonably well, many fail under the MTLR survival formulation due to the reversed logit dynamics.

To address this issue, we proposed HazardDev, a simple yet effective metric that leverages hazard deviation to capture distributional discrepancies specific to survival modeling. HazardDev consistently outperforms existing post-hoc OOD methods across all survival tasks, establishing a strong baseline for future research.

Future work can proceed along two directions. First, our results highlight that existing OOD detection approaches remain suboptimal for survival prediction, calling for new, survival-aware OOD detection algorithms that better capture uncertainty under covariate shifts. Second, the persistent gap between ID and OOD event prediction performance underscores the need to enhance model robustness and generalization. Improving the stability of event prediction under unseen acquisition settings will be crucial for deploying reliable and trustworthy survival models in real clinical scenarios.
\else
    \section{Conclusion and Discussion}


We introduced CURE-OOD, to our knowledge, the first benchmark for OOD detection in cancer survival prediction built on real-world clinical CT data, through a controlled evaluation protocol that uses acquisition parameter variation to instantiate realistic covariate shifts. Through experiments spanning multiple acquisition shifts and representative OOD detectors, we demonstrate that acquisition shifts substantially degrade survival prediction, and that OOD methods developed for classification can fail in survival prediction due to distinct logit dynamics.
As a simple survival-aware reference baseline, HazardDev further suggests that hazard-based signals can provide a useful direction for OOD detection in survival prediction. Future work may build on CURE-OOD to develop more task-aligned OOD detection and robustness evaluation under distribution shifts.

\fi

{
    \small
    \bibliographystyle{tmlr}
    \bibliography{main}
}

\clearpage
\appendix
\section*{Appendix}

\begin{figure*}[!t]
    \centering
    \begin{subfigure}[b]{1\linewidth}
        \centering
        \includegraphics[width=\linewidth]{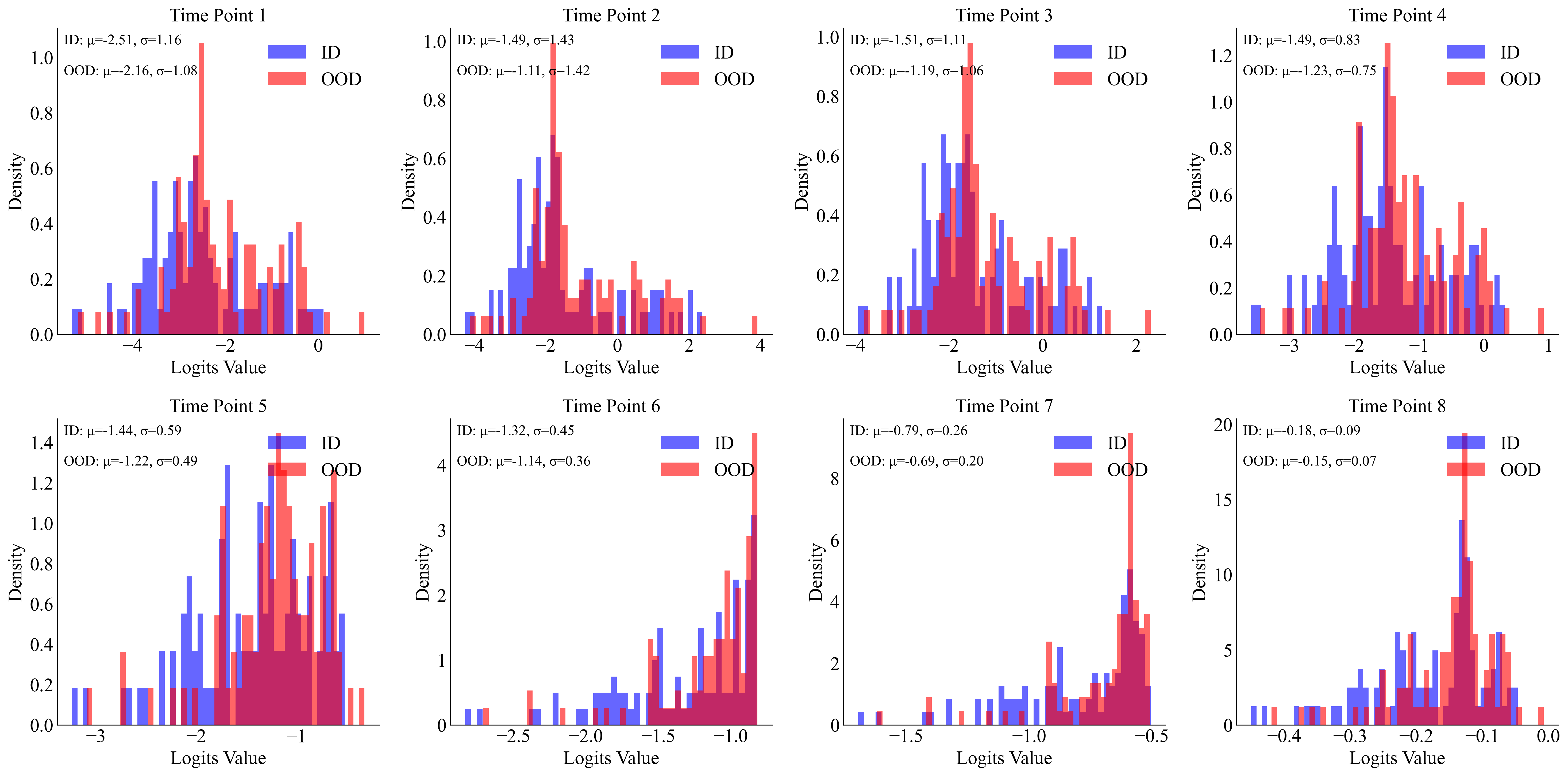}
        \caption{Pixel Spacing}
        \label{fig:os_logits_pixel}
    \end{subfigure}
    \vspace{30pt}

    \begin{subfigure}[b]{1\linewidth}
        \centering
        \includegraphics[width=\linewidth]{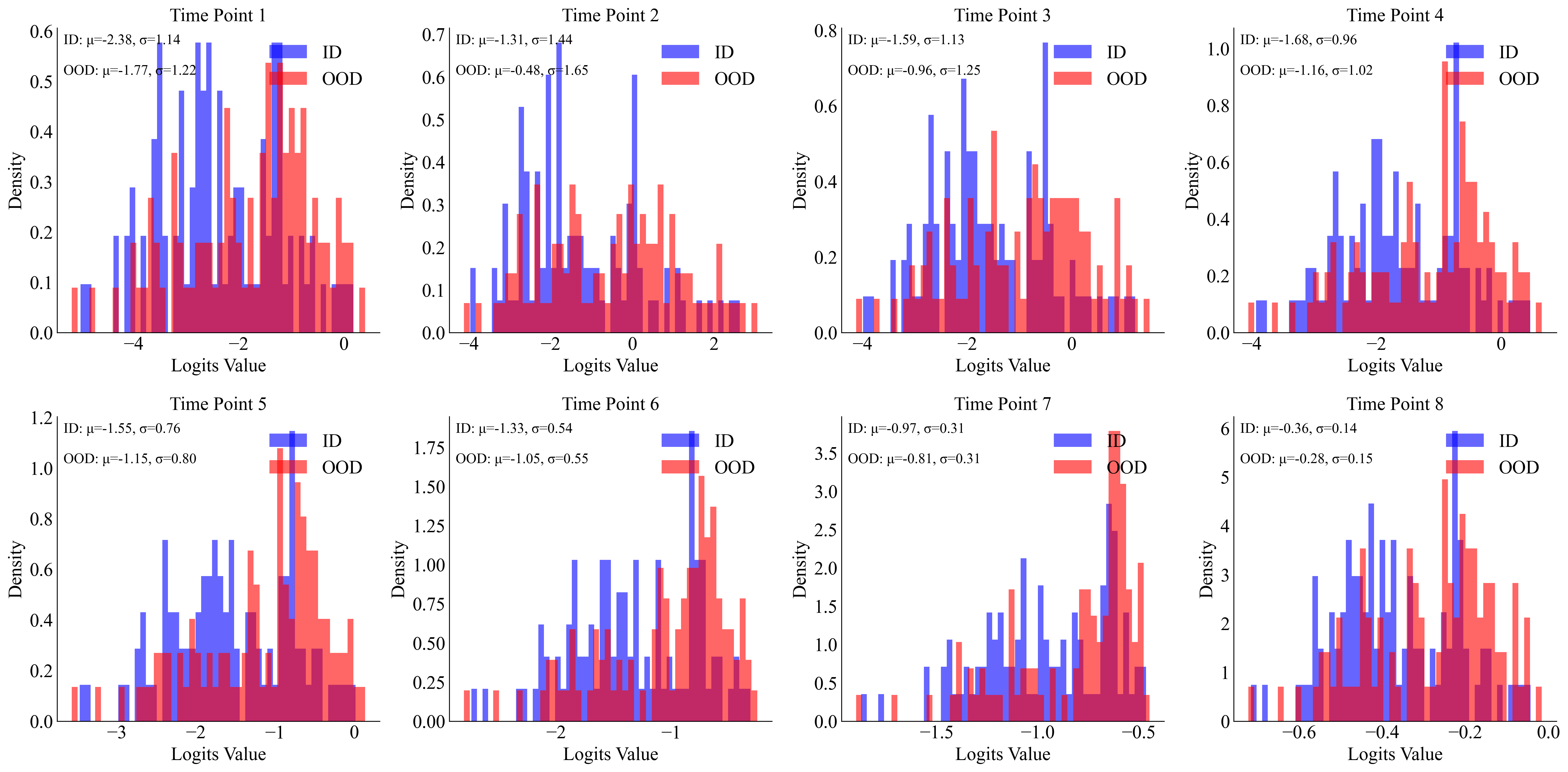}
        \caption{Exposure Time}
        \label{fig:os_logits_expo}
    \end{subfigure}
    \vskip 3pt
    \caption{
    Visualization of MTLR logits distributions on the OS task under two acquisition shifts: 
    (a) pixel spacing and (b) exposure time. 
    For both shifts, ID samples tend to produce logits concentrated at lower values, 
    whereas OOD samples show a clear tendency toward larger logits. 
    This shift in the logits distribution indicates that acquisition differences systematically affect the model’s output confidence, 
    providing a useful signal for OOD detection.
    }
    \label{fig:os_logits_distribution}
\end{figure*}

\begin{figure*}[t]
  \centering
  \begin{subfigure}{0.495\linewidth}
    \includegraphics[width=\linewidth]{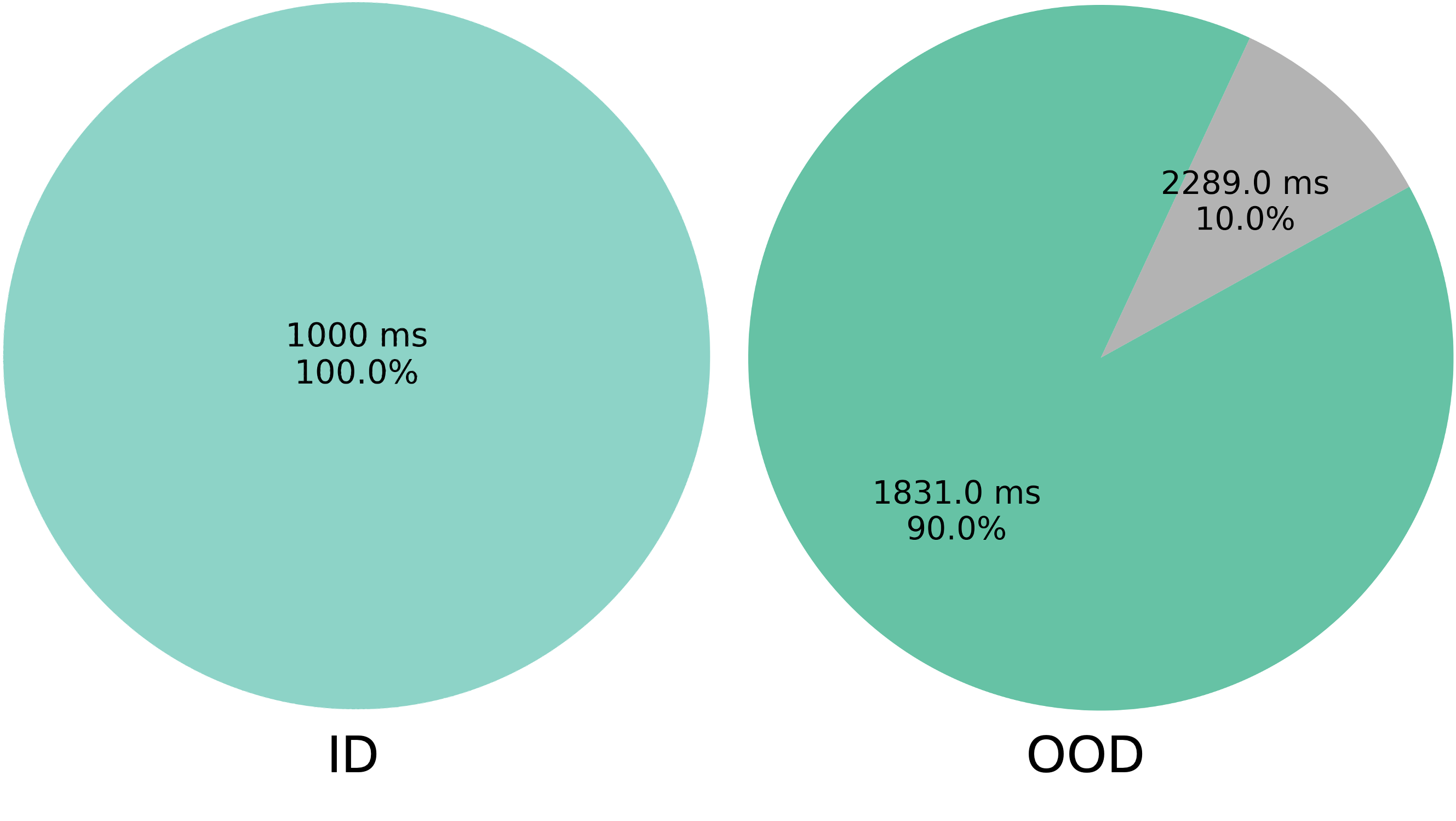}
    \caption{Exposure time distribution across ID and OOD.}
    \label{fig:expo_distribution}
  \end{subfigure}\hfill
  \begin{subfigure}{0.495\linewidth}
    \includegraphics[width=\linewidth]{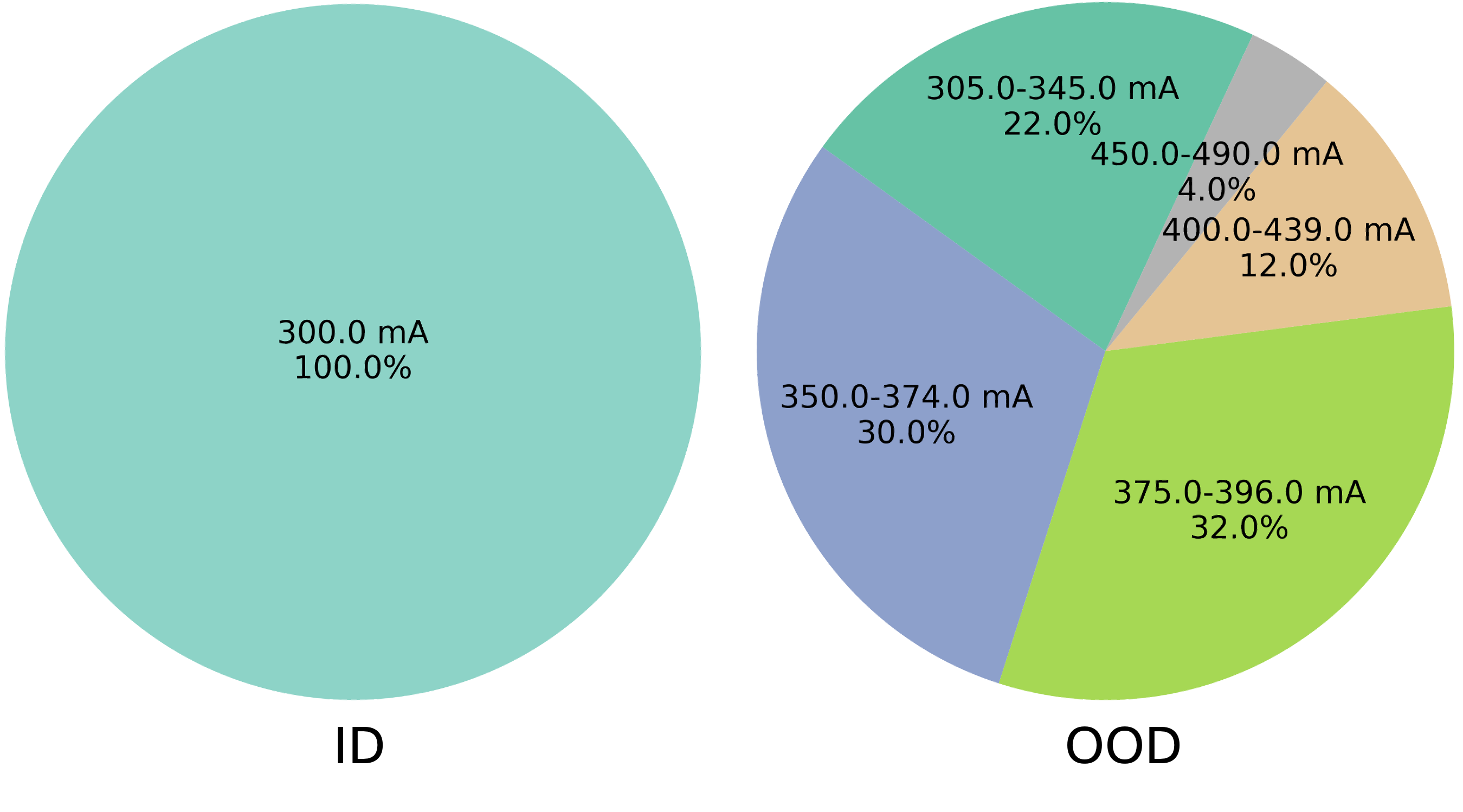}
    \caption{X-ray tube current distribution across ID and OOD.}
    \label{fig:xray_distribution}
  \end{subfigure}
  \caption{Distributions of key acquisition parameters used to construct ID and OOD domains in the CURE-OOD benchmark. Each parameter exhibits distinct value ranges.}
  \label{fig:datadistribution}
\end{figure*}

\section{Gradient Dynamics of MTLR and Logit Behavior}
\label{sec:reverse}

To understand why logits-based OOD detection methods such as Energy and SCALE behave inversely on survival modeling tasks, we analyze the formulation of the MTLR model.

MTLR divides the continuous survival time into $m$ discrete intervals $\{t_1, t_2, \dots, t_m\}$ and models the conditional probability of survival at each interval. 
For a patient with feature $\mathbf{x}$ and survival status sequence $y = (y_1, y_2, \dots, y_m)$, where $y_i = 0$ indicates survival up to $t_i$, the likelihood is defined as:
\begin{align}
P_{\Theta}(Y = y \mid \mathbf{x}) 
&= 
\frac{
\exp\!\left( \sum_{i=1}^{m} y_i(\mathbf{\theta}_i^\top \mathbf{x} + b_i) \right)
}{
\sum_{k=0}^{m} \exp(f_{\Theta}(\mathbf{x}, k))
},
\label{eq:mtlr_likelihood}
\\[3pt]
&\text{where } f_{\Theta}(\mathbf{x}, k) = \sum_{i=k+1}^{m} (\mathbf{\theta}_i^\top \mathbf{x} + b_i).
\notag
\end{align}

Here, $f_{\Theta}(\mathbf{x}, k)$ represents the unnormalized logit (score) associated with the event occurring at time interval $k$.
The denominator normalizes these scores over all possible event times, forming a valid probability distribution across discrete survival intervals.

The model is optimized by minimizing the \textit{regularized negative log-likelihood}, 
which corresponds to maximizing the log-likelihood of the observed survival sequences 
with additional $\ell_2$ and smoothness regularization terms~\citep{yu2011learning}. 
Formally, the optimization objective can be written as:
\begin{equation}
\min_{\Theta} 
\frac{C_1}{2} \sum_{j=1}^{m} \|\boldsymbol{\theta}_j\|_2^2
+ \frac{C_2}{2} \sum_{j=1}^{m-1} \|\boldsymbol{\theta}_{j+1} - \boldsymbol{\theta}_j\|_2^2
- \mathcal{L}_{\text{MTLR}},
\end{equation}
where $\mathcal{L}_{\text{MTLR}}$ denotes the log-likelihood term:
\begin{equation}
\begin{split}
\mathcal{L}_{\text{MTLR}} =
\sum_{i=1}^{N}
\Big[
& \sum_{j=1}^{m} y_j(s_i)(\mathbf{\theta}_j^\top \mathbf{x}_i + b_j) \\
& - \log \sum_{k=0}^{m} \exp f_{\Theta}(\mathbf{x}_i, k)
\Big].
\end{split}
\end{equation}

The first regularizer $\sum_j \|\boldsymbol{\theta}_j\|_2^2$ constrains the parameter magnitude 
to prevent overfitting, while the second term $\sum_j \|\boldsymbol{\theta}_{j+1} - \boldsymbol{\theta}_j\|_2^2$ 
enforces smoothness across consecutive time intervals.

During optimization, the first component of $\mathcal{L}_{\text{MTLR}}$ increases the scores 
for observed event intervals, whereas the second term normalizes the probabilities 
across all possible time sequences. 
This formulation naturally drives the logits to form a temporal ordering. 
At early time intervals, where most samples remain alive, the logits become smaller (more negative).  
At later intervals, where events are more likely to occur, the logits gradually increase (become less negative).  
This trend is consistently observed across datasets, as shown in Fig.~\ref{fig:os_logits_distribution},  
where the mean logits shift upward from early to late time bins.  
We analyze the gradient dynamics below to theoretically explain why this monotonic pattern emerges during training.

\subsection{Gradient Analysis and the Tendency Toward Monotonic Logits}
\label{app:mtlr-gradient}

We derive the gradient of the core likelihood term and show why optimization encourages
smaller (more negative) logits at early intervals and larger (less negative) logits at later intervals.

\paragraph{Setup.}
For a sample $(\mathbf{x}_i,s_i)$, define the normalizer
\[
S_i(\Theta) \coloneqq \sum_{k=0}^{m} \exp f_{\Theta}(\mathbf{x}_i,k), 
\qquad
\pi_{ik} \coloneqq \frac{\exp f_{\Theta}(\mathbf{x}_i,k)}{S_i(\Theta)} .
\]
Recall that
\( f_{\Theta}(\mathbf{x},k)=\sum_{r=k+1}^{m}(\boldsymbol{\theta}_r^\top \mathbf{x}+b_r) \),
so for a fixed \(j\in\{1,\dots,m\}\),
\[
\frac{\partial f_{\Theta}(\mathbf{x}_i,k)}{\partial \boldsymbol{\theta}_j}
=
\begin{cases}
\mathbf{x}_i, & k<j,\\
\mathbf{0}, & k\ge j.
\end{cases}
\tag{A.1}
\label{eq:df-dtheta}
\]

\paragraph{Gradient of the per-sample objective.}
For the core likelihood term
\[
\ell_i(\Theta)=
\sum_{r=1}^{m} y_r(s_i)\,(\boldsymbol{\theta}_r^\top \mathbf{x}_i+b_r)
-\log S_i(\Theta),
\]
the gradient with respect to $\boldsymbol{\theta}_j$ is
\begin{align}
\frac{\partial \ell_i(\Theta)}{\partial \boldsymbol{\theta}_j}
&=
y_j(s_i)\,\mathbf{x}_i
-
\frac{\partial}{\partial \boldsymbol{\theta}_j}\log S_i(\Theta).
\tag{A.2}
\label{eq:grad-li-partial}
\end{align}
The first term directly increases the score for interval $j$ when the sample survives past $t_j$.  
The second term involves the derivative of the log-normalizer, which we compute next.

\paragraph{Derivative of the log-normalizer.}
Using \eqref{eq:df-dtheta}, we have
\begin{align}
\frac{\partial}{\partial \boldsymbol{\theta}_j}\log S_i(\Theta)
&=
\frac{1}{S_i(\Theta)}
\sum_{k=0}^{m} \exp f_{\Theta}(\mathbf{x}_i,k)\;
\frac{\partial f_{\Theta}(\mathbf{x}_i,k)}{\partial \boldsymbol{\theta}_j}
\notag\\
&=
\frac{1}{S_i(\Theta)}
\sum_{k=0}^{j-1} \exp f_{\Theta}(\mathbf{x}_i,k)\;\mathbf{x}_i
=
\sum_{k=0}^{j-1} \pi_{ik}\,\mathbf{x}_i .
\tag{A.3}
\label{eq:dlogZ}
\end{align}

Substituting \eqref{eq:dlogZ} into \eqref{eq:grad-li-partial} yields
\begin{align}
\frac{\partial \ell_i(\Theta)}{\partial \boldsymbol{\theta}_j}
&=
\bigg(
y_j(s_i)
-
\sum_{k=0}^{j-1}\pi_{ik}
\bigg)\mathbf{x}_i .
\tag{A.4}
\label{eq:per-sample-grad}
\end{align}
Summing over all samples gives the overall gradient
\begin{equation}
\frac{\partial \mathcal{L}_{\text{MTLR}}}{\partial \boldsymbol{\theta}_j}
=
\sum_{i=1}^{N}
\bigg(
y_j(s_i)-\sum_{k=0}^{j-1}\pi_{ik}
\bigg)\mathbf{x}_i ,
\tag{A.5}
\label{eq:full-grad}
\end{equation}
to which one may add the $\ell_2$ and smoothness regularizers from~\citep{yu2011learning} if included.

\paragraph{Interpretation.}
The term \(y_j(s_i)\) indicates whether sample \(i\) has \emph{survived past} \(t_j\) (with the 0/1 encoding in the main text).
The cumulative softmax mass \(\sum_{k=0}^{j-1}\pi_{ik}\) is the model's probability assigned to the event occurring
\emph{before} \(t_j\). Hence the gradient \eqref{eq:full-grad} pushes the parameters so that:
\begin{itemize}
\item if many samples survive past \(t_j\) (empirically common for early \(j\)), then \(y_j(s_i)\) tends to dominate,
and the update decreases \(\sum_{k<j}\pi_{ik}\) by \emph{lowering} the logits \(f_{\Theta}(\mathbf{x}_i,k)\) for \(k<j\)
(via \eqref{eq:df-dtheta}), making early-interval logits more negative;
\item conversely, for later \(j\) where events are more likely, the model is encouraged to allocate
more mass to later intervals, effectively making those logits less negative.
\end{itemize}
Aggregated over the dataset, this drives a characteristic pattern in which early-interval logits become smaller
(more negative) while later-interval logits become larger (less negative), so that the learned logits
\emph{tend to} exhibit a monotonic progression across time, consistent with increasing event risk.
We provide a detailed derivation of this mechanism in Section~\ref{app:early-mass-decrease}, 
showing how each optimization step explicitly suppresses the cumulative probability of early intervals.

\subsection{From the Gradient to Early-Interval Probability Suppression}
\label{app:early-mass-decrease}
Building upon the gradient expression above, we further examine how each optimization
step explicitly suppresses the early cumulative probability mass $\sum_{k<j}\pi_{ik}$,
leading to a systematic decrease in early logits.

We show why maximizing the likelihood drives down the cumulative early probability
$\sum_{k<j}\pi_{ik}$ and thus pushes early logits $f_{\Theta}(\mathbf{x}_i,k)$, $k<j$, to more
negative values.

\paragraph{SGD update on $\boldsymbol{\theta}_j$.}
Consider one SGD step on sample $(\mathbf{x}_i,s_i)$ with stepsize $\eta>0$:
\[
\Delta\boldsymbol{\theta}_j
=\eta\,
\frac{\partial \ell_i}{\partial \boldsymbol{\theta}_j}
=
\eta\Big(y_j(s_i)-\sum_{k=0}^{j-1}\pi_{ik}\Big)\mathbf{x}_i,
\tag{A.5}
\label{eq:sgd-theta}
\]
where $\pi_{ik}=\exp f_{\Theta}(\mathbf{x}_i,k)/\sum_{r}\exp f_{\Theta}(\mathbf{x}_i,r)$.


By \eqref{eq:df-dtheta}, a small parameter update $\Delta\boldsymbol{\theta}_j$ induces a corresponding change in the logits $f_{\Theta}(\mathbf{x}_i,k)$ through the chain rule:
\begin{align}
\Delta f_{\Theta}(\mathbf{x}_i,k)
&\approx
\Big\langle
\nabla_{\boldsymbol{\theta}_j} f_{\Theta}(\mathbf{x}_i,k),
\Delta\boldsymbol{\theta}_j
\Big\rangle
=
\Big(\tfrac{\partial f_{\Theta}(\mathbf{x}_i,k)}{\partial \boldsymbol{\theta}_j}\Big)^{\!\top}
\Delta\boldsymbol{\theta}_j
\notag\\
&=
\mathbf{x}_i^{\top}\Delta\boldsymbol{\theta}_j,
\qquad k<j,
\tag{A.6a}
\end{align}
since $\partial f_{\Theta}(\mathbf{x}_i,k)/\partial\boldsymbol{\theta}_j=\mathbf{x}_i$ only holds for $k<j$.
Substituting the update rule from~\eqref{eq:sgd-theta} gives
\begin{align}
\Delta f_{\Theta}(\mathbf{x}_i,k)
&=
\mathbf{x}_i^{\top}
\Big[
\eta\Big(y_j(s_i)-\sum_{r=0}^{j-1}\pi_{ir}\Big)\mathbf{x}_i
\Big]
\notag\\
&=
\eta
\Big(
y_j(s_i)
-
\sum_{r=0}^{j-1}\pi_{ir}
\Big)
\|\mathbf{x}_i\|_2^{2},
\qquad k<j,
\tag{A.6}
\label{eq:delta-f}
\end{align}
and leaves $f_{\Theta}(\mathbf{x}_i,k)$ unchanged for $k\ge j$ since $\partial f_{\Theta}/\partial\boldsymbol{\theta}_j=\mathbf{0}$ in that case.

\paragraph{Sign of the update at early $j$.}
With the $0\!\to\!1$ encoding in MTLR, $y_j(s_i)=1$ iff the sample has \emph{survived past} $t_j$.
Empirically, at \emph{early} $j$ most samples survive past $t_j$, so for many $i$ we have
$y_j(s_i)\approx 1$ while the model still allocates nonzero mass to earlier events,
$\sum_{r<j}\pi_{ir}>0$.
Hence the coefficient in \eqref{eq:delta-f} is typically a positive number slightly less than~1,
\[
y_j(s_i)-\sum_{r<j}\pi_{ir}\ \approx\ 1-\epsilon,\quad \epsilon>0.
\]
Although this term itself is positive, its effect on earlier logits $f_{\Theta}(\mathbf{x}_i,k)$ with $k<j$
is \emph{negative}. 
This arises because each logit $f_{\Theta}(\mathbf{x}_i,k)$ accumulates all later parameters 
$\{\boldsymbol{\theta}_r : r>k\}$ in
$f_{\Theta}(\mathbf{x}_i,k)=\sum_{r=k+1}^{m}(\boldsymbol{\theta}_r^\top\mathbf{x}_i+b_r)$.
When $\boldsymbol{\theta}_j$ increases, it raises the logits of later intervals more strongly than those of earlier ones,
thereby reducing the relative magnitude of early logits under the softmax normalization.
Consequently,
\[
\Delta f_{\Theta}(\mathbf{x}_i,k)\le0,\qquad k<j,
\tag{A.7}
\label{eq:sign-delta-f}
\]
meaning that SGD decreases all early logits, pushing them toward more negative values.

\paragraph{Effect on the early cumulative probability.}
The softmax probabilities are monotone in their logits. Writing
$\boldsymbol{f}_i=(f_{\Theta}(\mathbf{x}_i,0),\dots,f_{\Theta}(\mathbf{x}_i,m))$ and
$\boldsymbol{\pi}_i=\operatorname{softmax}(\boldsymbol{f}_i)$, the Jacobian is
\[
\frac{\partial \pi_{iq}}{\partial f_{\Theta}(\mathbf{x}_i,k)}
=\pi_{iq}\big(\mathbb{1}\{q=k\}-\pi_{ik}\big).
\tag{A.8}
\]
If all early logits $\{f_{\Theta}(\mathbf{x}_i,k):k<j\}$ decrease by
$\{\Delta f_{\Theta}(\mathbf{x}_i,k)\le 0\}$ while the others stay fixed, then the total
early mass strictly decreases:
\[
\Delta\Big(\sum_{k<j}\pi_{ik}\Big)
=\sum_{k<j}\sum_{q=0}^{m}\frac{\partial \pi_{iq}}{\partial f_{\Theta}(\mathbf{x}_i,k)}\,
\Delta f_{\Theta}(\mathbf{x}_i,k)\ <\ 0.
\tag{A.9}
\]
Hence the update \eqref{eq:sgd-theta} \emph{suppresses} the cumulative early probability.

\paragraph{Population effect.}
Aggregating such SGD steps over the dataset, early intervals (where survival is prevalent)
receive repeated updates of the form \eqref{eq:sign-delta-f}, which consistently lower their
logits and probability mass; later intervals receive comparatively fewer such decreases.
Consequently, the learned logits tend to become smaller (more negative) at early $k$ and larger
(less negative) at late $k$, yielding the empirical monotonic trend reported in the main text.

\subsection{Why ID Logits Are More Negative}

For ID samples, the model has been optimized to align the logits with the empirical hazard trend of the training data. 
At early time points, the survival probability is high and the event likelihood is low, leading the logits $f_{\Theta}(\mathbf{x}, k)$ to take more negative values.
At later time points, as the risk of event increases, the logits gradually become less negative. 
This pattern is a direct consequence of likelihood maximization and the monotonicity constraint imposed by the MTLR framework.

\begin{figure}[!t]
    \centering
    \begin{subfigure}[b]{0.48\linewidth}
        \centering
        \includegraphics[width=\linewidth]{img/OS_average_hazard_curves_expo.png}
        \caption{Exposure Time}
        \label{fig:hazard_expo_suppl}
    \end{subfigure}
    \hfill
    \begin{subfigure}[b]{0.48\linewidth}
        \centering
        \includegraphics[width=\linewidth]{img/OS_average_hazard_curves_pixel.png}
        \caption{Pixel Spacing}
        \label{fig:hazard_pixel_suppl}
    \end{subfigure}

    \vskip 3pt 

    \begin{subfigure}[b]{0.48\linewidth}
        \centering
        \includegraphics[width=\linewidth]{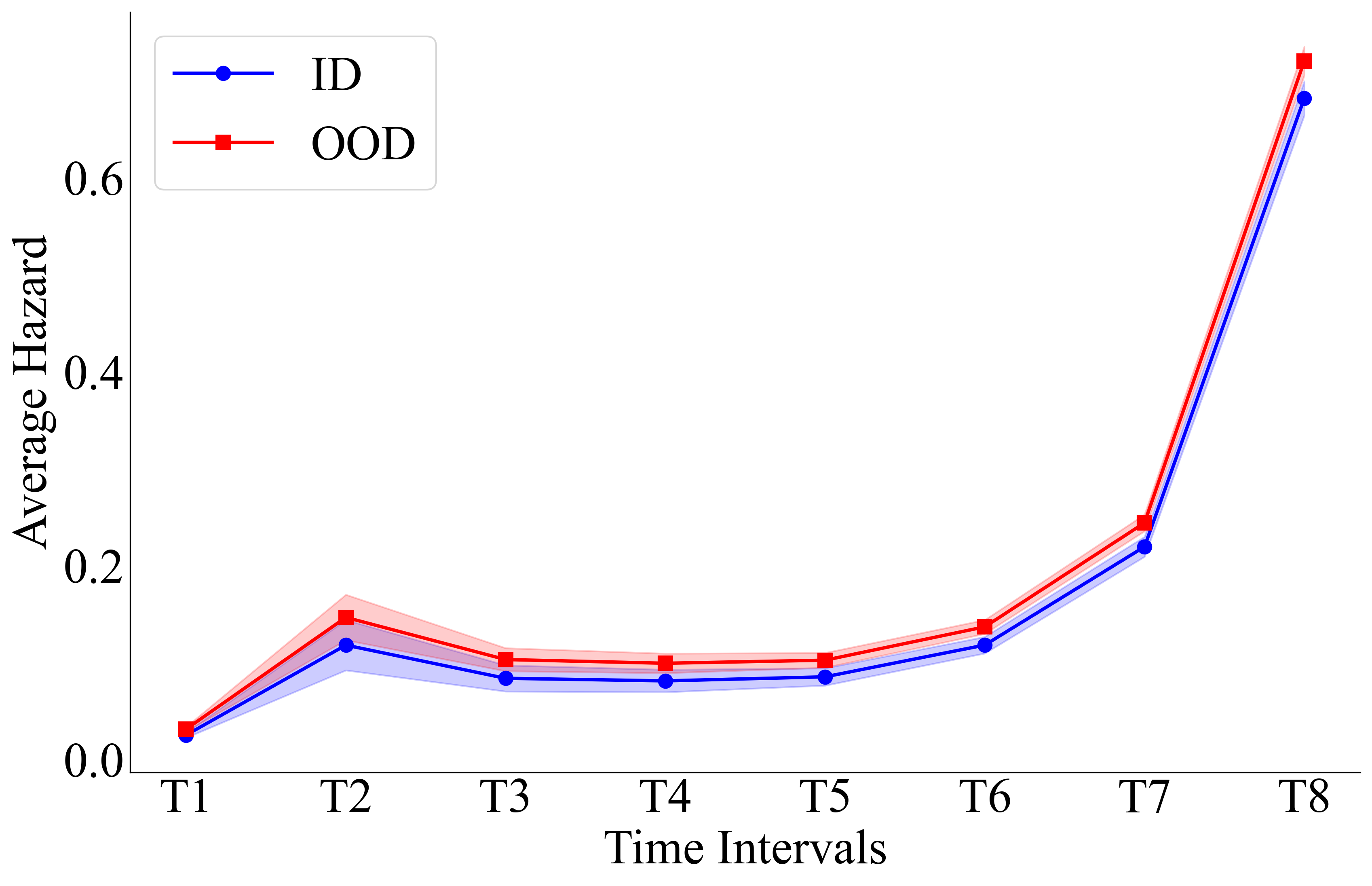}
        \caption{Slice Thickness}
        \label{fig:hazard_thickness}
    \end{subfigure}
    \hfill
    \begin{subfigure}[b]{0.48\linewidth}
        \centering
        \includegraphics[width=\linewidth]{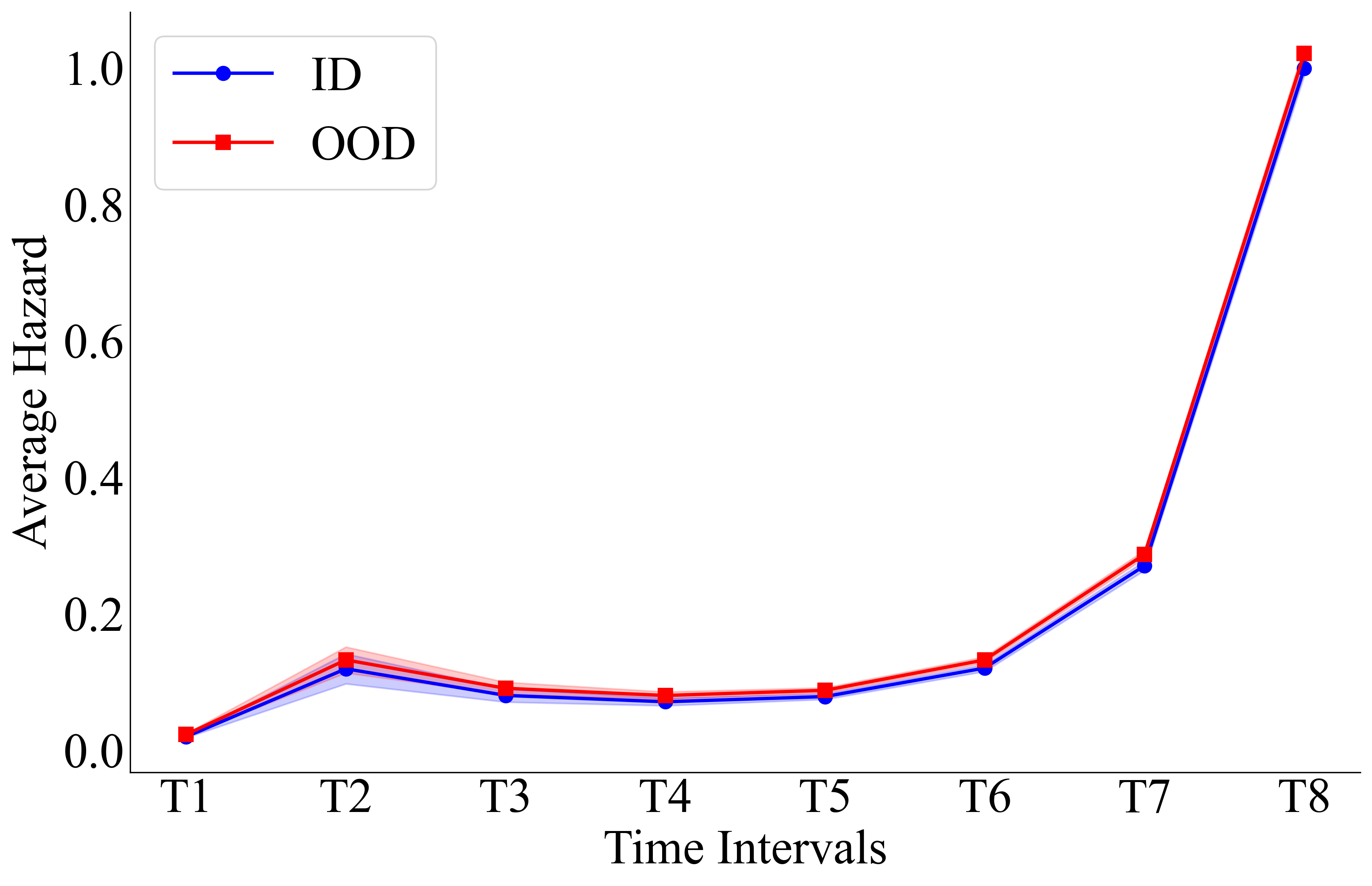}
        \caption{X-Ray Tube Current}
        \label{fig:hazard_xray}
    \end{subfigure}

    \caption{
Mean hazard curves of ID and OOD test sets on the OS task across four acquisition shifts: 
(a) exposure time, (b) pixel spacing, (c) slice thickness, and (d) X-ray tube current. 
The shaded areas denote one standard deviation, showing the variability of predicted hazard values across samples. 
Across all shifts, ID samples consistently exhibit higher mean hazard values than OOD samples, indicating that hazard magnitude can serve as an effective signal for OOD detection.
}
    \label{fig:hazard_curves}
\end{figure}

\subsection{Why OOD Logits Become Larger}

When encountering OOD samples, the feature representations $\mathbf{x}$ deviate from the training distribution. 
In these unseen regions, the learned parameters $\{\mathbf{\theta}_i, b_i\}$ produce less calibrated responses, and the model tends to generate logits that are closer to zero. 
Formally, since $f_{\Theta}(\mathbf{x}, k)$ is a linear projection of $\mathbf{x}$,
\[
f_{\Theta}(\mathbf{x}, k) = \sum_{i=k+1}^{m} (\mathbf{\theta}_i^\top \mathbf{x} + b_i),
\]
a shift in $\mathbf{x}$ toward regions unobserved during training leads to a reduction in the projection magnitude $|\mathbf{\theta}_i^\top \mathbf{x}|$, effectively making logits less negative. 
This reflects increased predictive uncertainty and weaker separation between survival intervals.

\subsection{Consequence for Logit-Based OOD Detection}
Most classification-based OOD detection methods, such as Energy or SCALE, assume that OOD samples produce logits with smaller magnitudes (lower confidence) than ID samples. 
However, in MTLR, the opposite occurs: ID samples yield strongly negative logits due to their alignment with the high-survival regions of the training data, while OOD samples result in less negative logits. 
Consequently, the conventional logit-based energy scores are inverted, causing ID samples to be misidentified as OOD and vice versa.

This inversion originates from the interplay between the monotonic survival constraint and distributional shifts in $\mathbf{x}$. 
Empirically, we observe this consistent polarity reversal across most shifts in the CURE-OOD benchmark.

\end{document}